# Is attention all you need in medical image analysis? A review.

Giorgos Papanastasiou, Nikolaos Dikaios, Jiahao Huang, Chengjia Wang, and Guang Yang.

***Abstract*— Medical imaging is a key component in clinical diagnosis, treatment planning and clinical trial design, accounting for almost 90% of all healthcare data. CNNs achieved performance gains in medical image analysis (MIA) over the last years. CNNs can efficiently model local pixel interactions and be trained on small-scale MI data. The main disadvantage of typical CNN models is that they ignore global pixel relationships within images, which limits their generalisation ability to understand out-of-distribution data with different "global" information. The recent progress of Artificial Intelligence gave rise to Transformers, which can learn global relationships from data. However, full Transformer models need to be trained on large-scale data and involve tremendous computational complexity. Attention and Transformer compartments ("Transf/Attention") which can well maintain properties for modelling global relationships, have been proposed as lighter alternatives of full Transformers. Recently, there is an increasing trend to co-pollinate complementary local-global properties from CNN and Transf/Attention architectures, which led to a new era of hybrid models. The past years have witnessed substantial growth in hybrid CNN-Transf/Attention models across diverse MIA problems. In this systematic review, we survey existing hybrid CNN-Transf/Attention models, review and unravel key architectural designs, analyse breakthroughs, and evaluate current and future opportunities as well as challenges. We also introduced a comprehensive analysis framework on generalisation opportunities of scientific and clinical impact, based on which new data-driven domain generalisation and adaptation methods can be stimulated.**

## I. INTRODUCTION

### A. Medical image analysis and convolutions

Medical imaging (MI) is a key component in clinical diagnosis, treatment planning, and clinical trial design, accounting for almost 90% of all healthcare data [1, 2]. Medical image analysis (MIA)-derived imaging biomarkers can improve early disease diagnosis, therapy design and treatment response monitoring, beyond visual radiology assessments. MIA is an important component of clinical research, innovation and application [3-6].

The European Society of Radiology in coordination with the Radiological Society of North America, has recently provided recommendations for clinically validating MIA techniques [3-5]. In the era of rapid artificial intelligence (AI) developments and to establish the clinical translation of AI, it is important to review and develop guidelines for innovative AI models.

Since the first "deep" convolutional neural network (CNN) developed by Krizhevsky et al. in 2012 which outperformed the previous state-of-the-art (SOTA, non-deep learning) algorithms on the ImageNet dataset [7], CNNs demonstrated numerous performance gains across all MIA tasks: segmentation, reconstruction, synthesis, denoising, registration, classification, and pathology detection [2]. However, typical CNNs focus on modelling information through small convolutional filter footprints and shared weights, which comes at the cost of introducing local receptive fields thus, limiting their ability to directly model long-range (global) pixel interactions within images. Hence, despite their important advances, CNN-based networks are still focusing on local-scale modelling, with low generic "local-global" modelling capabilities. Their limited ability to model both local and global information from images adds barriers to model generalisability (e.g. across MIA domains or pathology settings) and transfer learning (from one MI modality to another) properties of pure CNN models [2].

### B. Hybridisation with attention convolutions

First introduced by Bahdanau et al. in 2014, the attention mechanism was initially designed to learn long-range dependencies in natural language processing and improve machine translation [8]. The attention mechanism allows to (soft-)search for a set of positions in a source sentence where the most relevant information is concentrated, encouraging the model to predict a target word based on the context vectors associated with these source positions and all the previous generated target words [8]. Following attention, the development of the self-attention mechanism in 2016 was designed so that each position (building block) within a self-attention layer (known as query, key and value) can attend to all positions in the output of the previous layer [9-10], as an additional technique to enhance modelling of long-range dependencies.

The introduction of self-attention and attention mechanisms in the Transformer models made it possible to increase the receptive field and thus, became an efficient solution for modelling long-range dependencies from images [10-12], with promising results in the field of MIA [13-16]. The Vision Transformer (ViT) models recruit consecutive multi-head self-attention and attention mechanisms in image patches and have been suggested to even fully replace pure CNN models [10]. The basic concept in ViT is to convert input images to a series of image patches which in turn are transformed into vectors and

This study has been partially supported by the MIS (5154714) of the National Recovery and Resilience Plan Greece 2.0 funded by EU under the NextGenerationEU program, ERC IMI (101005122), H2020 (952172), MRC (MC/PC/21013), Royal Society (IEC\NSFC\211235), and UKRI Future Leaders Fellowship (MR/V023799/1). GP is with the Archimedes Unit, Athena Research Centre, Athens, 15125, Greece (g.papanastasiou@essex.ac.uk). ND is with Mathematics Research Centre, Academy of Athens, Athens 10679, Greece (ndikaios@Academyofathens.gr). CW is with the School of Mathematical and Computer Sciences, Heriot Watt, Edinburgh EH14 4AS, UK (Chengjia.Wang@hw.ac.uk). GY and JH are with the Bioengineering Department and Imperial-X, Imperial College London, London W12 7SL, UK (g.yang@imperial.ac.uk; j.huang21@imperial.ac.uk).

can be represented as "words" in a normal Transformer. However, as the relationships between an image patch and all other image patches are computed, the computational complexity of the multi-head self-attention modules in ViT becomes quadratic to image size, adding substantial challenges in the setting of analysing high spatial resolution images. Swin Transformers (ST) were designed to overcome these challenges by performing self-attention in non-overlapped image patches [11, 12]. Despite this, ST need to consecutively learn a stack of two successive self-attention blocks with regular and shifted windowing configurations, respectively. This adds computational complexity and limits their applicability in MIA tasks such as segmentation, pathology detection, denoising, reconstruction and registration, where dense predictions at the pixel level and learning representations from high content images are necessary. This is one of the main reasons why full ViT and ST models have been limited to medical image classification and object detection tasks [10-17].

To reduce the computational complexity and to address both local and global learning in MIA, self-attention and Transformer blocks were incorporated into CNN model architectures (thereafter called as "hybrid" CNN-Transf/Attention models), giving rise to hybrid models. Current evidence shows that by combining local and global modelling capabilities, these hybrid CNN-Transf/Attention models consistently outperform previous SOTA techniques across different MIA tasks [13-21]. Hybrid models can potentially be also used to improve model interpretability [22-25].

However, the main drawback of these hybrid models is that they are enormously complex as they have been developed to address particular problems in MIA, which means that their domain generalizability (e.g. from CT to MRI, or from lung to cardiac applications) and transfer learning capabilities can be challenging processes. Given their substantial growth, it is important to methodically assess whether these techniques can generalise across imaging modalities, MIA tasks and clinical applications, or may be over-engineered to specific MIA problems.

In this work, we review the evolution of the hybrid models for in vivo MI: magnetic resonance imaging (MRI), computed tomography (CT), positron emission tomography (PET), ultrasound, x-Rays and retinal imaging. There are numerous recent surveys that describe technical details of CNN models and how these were used to address specific needs in MI [2, 26-29], as well as some recent survey on ViT in MI [17, 30-33]. Differing from previous reviews, we developed a comprehensive systematic review based on the Preferred Reporting Items for Systematic Reviews and Meta-Analyses (PRISMA) guidelines for hybrid CNN-Transf/Attention models in MI. We categorised published work on hybrid CNN-Transf/Attention models in MI, analysed key architectural designs and quantitatively as well as qualitatively unravelled the evolution of CNN-Transf/Attention models.

To improve clarity and understanding on these novel techniques, we critically review whether such hybrid models outperform their pure CNN counterparts. We review technical and computational complexities and discuss domain generalization strategies, based on the MI modality, downstream task, and clinical application. We focus on unravelling the importance and potential drawbacks of hybrid models. Finally, we discuss opportunities, challenges (with mitigations, where applicable) and future perspectives of the post-hybrid model era. We consider these review concepts as important pathways towards harmonising and translating these novel techniques into clinically meaningful MIA.

## II. METHODS

### A. Literature review strategy

We performed a systematic review of CNN-Transf/Attention models in MI published between January 1, 2019 and July 1, 2022 using Scopus, Web of Science and Pubmed, based on the PRISMA framework [34]. In our review, we refer to all hybrid models that involve any CNN and Transformer modules-including adaptations of self-attention and attention mechanisms, as hybrid CNN-Transf/Attention models. We only considered MI modalities that involve in vivo body imaging and thus, excluded microscopy and digital pathology slide imaging studies. Therefore, we focused on MRI, CT, PET-CT, ultrasound, retinal imaging and x-Rays.

Initial filtering: To broaden the research, we initially mined all publications by searching the following keywords in the abstract, title, and manuscript keywords: (transformer OR self-attention) AND (deep AND learning) OR (convolutional AND neural AND network). This led to 5,222 papers (see PRISMA flow in Fig. 1a). Subsequently, we focused the search by considering all different combinations of relevant keywords in the abstract, title and keywords of each paper, as follows: (transformer OR self-attention ) AND (deep AND learning ) OR (convolutional AND neural AND network) AND (medical AND imaging) OR (magnetic AND resonance AND imaging) OR (MRI) OR (computed AND tomography) OR (CT) OR (ultrasound) OR (positron AND emission AND tomography) OR TITLE-ABS-KEY (retin) OR TITLE-ABS-KEY (x-ray) OR TITLE-ABS-KEY (ray). By adding these terms, we removed all irrelevant to MI papers, which led to 656 papers from all three digital libraries. By excluding conference, review and archived (non-peer-reviewed) papers, we then removed all non-journal publications, leaving 352 journal papers for subsequent analysis.

Title and abstract screening: All authors screened titles and abstracts across all 352 journal peer-reviewed papers and removed all irrelevant to the field of study papers, leaving 128 papers for full text review.

Full text screening: Following full paper review, the authors removed 16 journal papers (14 non-relevant to MI or hybrid model studies and 2 papers not written in English). In total, 112 journal papers (thereafter, referred to as "articles") were included in our review analysis. See also data extraction for paper content that was reviewed.

### B. Data extraction

During evaluation of article full texts, we considered the following aspects: (1) year of publication; (2) MI modality; (3) CNN backbone; (4) Transf/Attention including all different attention subtypes; (5) MIA task (segmentation, reconstruction, synthesis, denoising, registration, classification, pathology detection); (6) organ or physiology system investigated/ imaged; (7) use of public or private data; (8) data augmentation technique used; implementation details: (9) model optimizer, (10) loss function, (11) metric used to evaluate the results and (12) size of training and testing data. We also considered (13) if computation expense (total number of parameters) was calculated and (14) whether performance was improved against non-hybrid baseline methods.

Furthermore, we assessed the articles in terms of generalisability following 2 objective criteria: whether a CNN-Transf/Attention architecture was a) trained and/ or evaluated on large unseen testing data, b) analysed data from heterogeneous modalities (e.g., different MRI or CT sequences, or MRI and CT, etc.) and/ or multi-modal analysis (image and text, images and genetics) and/ or was implemented in more than one organ area (e.g., brain and heart). Further, we identify challenges, opportunities and future trends that can be used as suggestions for future work in this field.

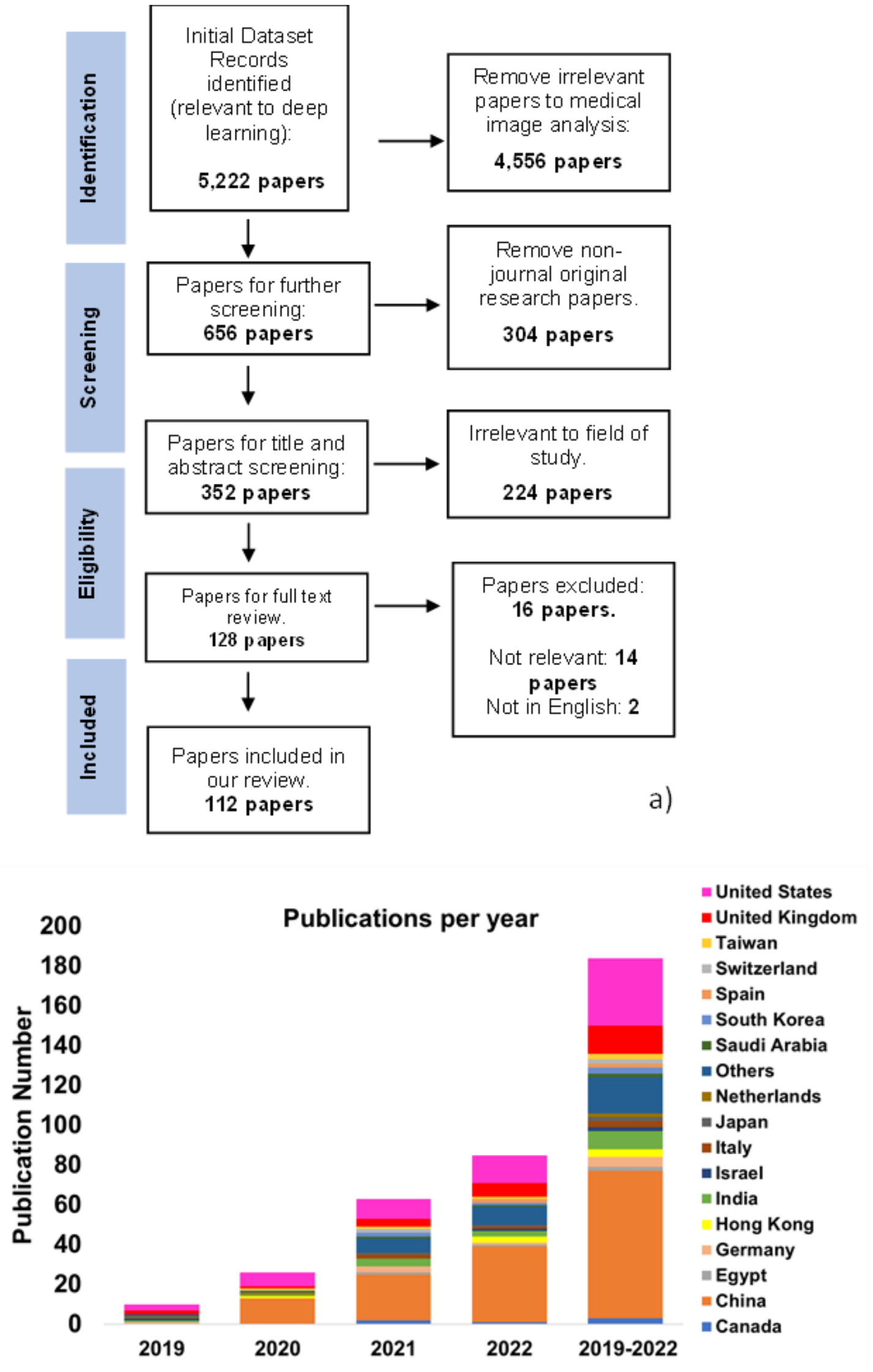

**Fig. 1a.** PRISMA flowchart. The flowchart illustrates inclusion and exclusion of papers at each review stage. **b.** Publications per year across the top 18 countries (in terms of publications record). Publications with affiliations from multiple countries have been accumulated on a per country basis.

## III. RESULTS

### A. Research trends

We studied published work on hybrid CNN-Transf/Attention models in MI and observed a consistent increase of these models in 2021 and 2022, against the first 2 years of our observation window.

In Fig. 1a, we present the PRISMA flow used to search and review articles. In Fig. 1b, we initially measured the country origin as derived from each affiliation across all articles (all affiliations were considered across publications). Considering the entire review period (2019-2022), the first ten countries in were: China (74 publications), USA (34), UK (14), India (9), Germany (5), Hong Kong (4), Canada (3), Taiwan (3), South Korea (3) and Italy (3).

Table I demonstrates all the articles grouped based on the MI modality, CNN backbone, Transf/Attention model and clinical application (organ) [13-21, 35-139]. Implementation details about the data augmentation technique, optimizer, loss function and the metrics used to evaluate the performance of each hybrid model across studies, are presented in Table II. Table III presents whether public and/ or private were analysed and information about the data size.

### B. Experimental settings and key architectural designs

#### Medical imaging modality recruited

We reviewed the publication record of the MI modality used per year (Fig. 2a). Most of the studies involved MRI (50 studies), followed by CT (42), retinal imaging (14), x-Rays (12), ultrasound (7) and PET-CT (5). Although MRI was less frequent than CT the first 2 years of our observation time frame, it outnumbered CT in the last two years (2021 and 2022).

#### CNN model used

In Fig. 2b, we demonstrate all CNN backbone models used across studies. Standard CNN architectures have been implemented in most of the studies (40 articles), followed by UNet (30), GAN (14), ResNet (14), DenseNet (7), None (i.e., no CNN backbone-only Transformer model used) (7), fully connected networks (FCN) (6) and VGG (3).

#### Transformer and attention mechanisms

Fig. 2c illustrates the evolution of Transf/Attention models recruited per year. It is obvious that self-attention mechanisms have been most widely used (64 studies out of 112 in total), followed by Transformer (22 studies), ViT (9 studies), channel- and spatial-attention (6), ST (4), attention (2) and other (11).

It is known that to exploit the performance capabilities of full Transformer models, a combination of large data and supercomputer facilities are necessary [10-12]. In our review, there were numerous studies that either analysed relatively small (i.e., <2,000 images) data (~27%), and/ or private data alone (~29%) and/ or did not report the data size (~21%). Details are presented in Table III. Furthermore, computational resources were not reported in most studies, with only 29 out of 83 reporting the number of model parameters and/ or time for training. Of note, only 8 out of 112 studies (~7%) described use of full original Transformer, ViT or ST models, with the rest ~93% involving: self-attention, channel- and spatial-attention, attention and simplified and light Transformer versions including transformer blocks (of stacked layers), layers, or encoders (Table I).

**Medical image analysis (downstream) task**

Further, we extracted all MIA (downstream) tasks across studies (Fig. 3a). Most of the studies aimed to solve segmentation tasks (43 studies), followed by pathology detection (39), classification (26), reconstruction (14), synthesis (10), denoising (3), localisation (2), registration (2) and radiology report generation (1). Note that pathology detection was mainly achieved as being the downstream task of joint segmentation [45, 57, 65, 116, 125], localisation [21] or classification [35-40, 48, 52, 63, 84-87, 89, 90, 93, 94, 97, 133], and in some cases was considered as a standalone (main) task [42-44, 46, 47, 49, 50, 54, 55, 92, 95, 96].

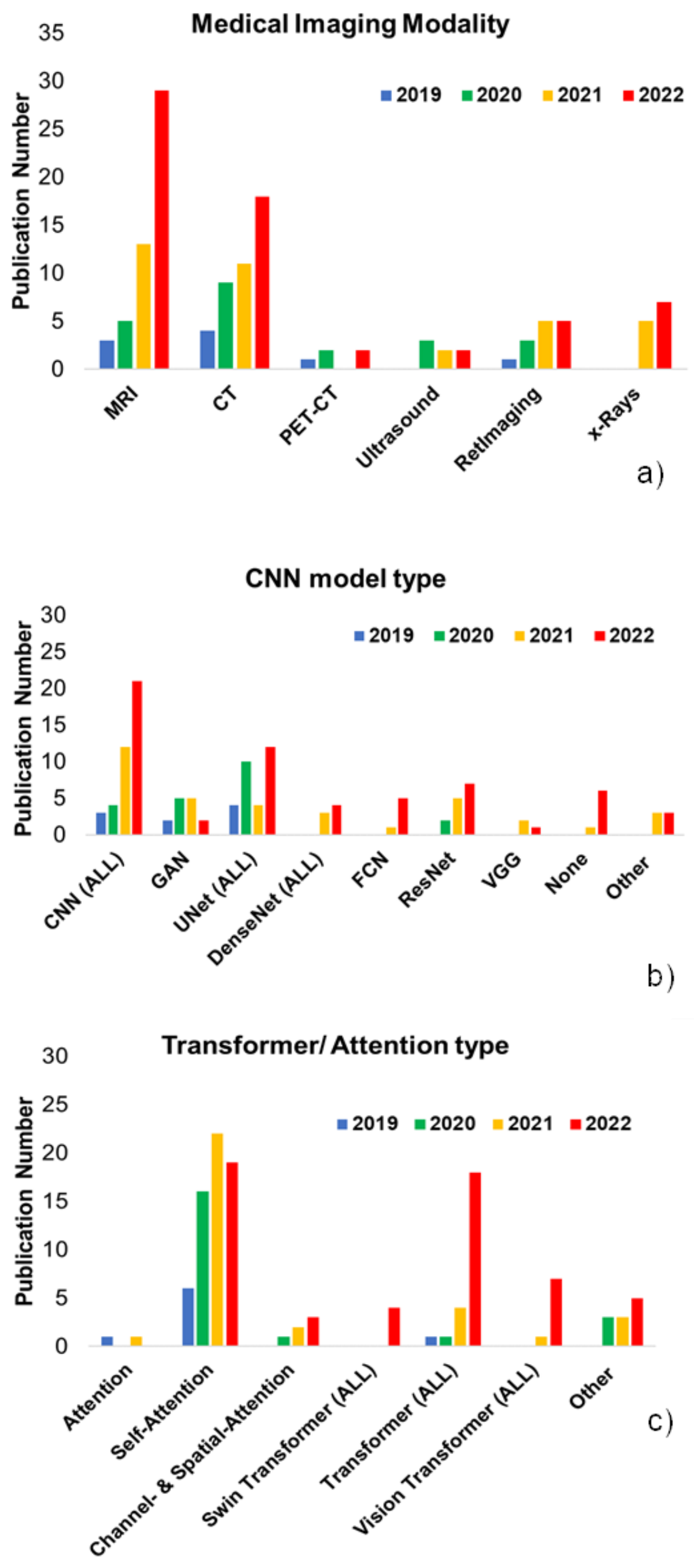

**Fig. 2.** Publication record over time for the Medical Imaging modality (a), the CNN model type (b), and Transf/Attention architectures (c). In b), the CNN (ALL) term describes all standard CNN models captured across studies: CNN encoder-decoder (E-D), CNN layers, CNN decoder only and CNN (E-D) (3D). GAN describes either GAN or CycleGAN models. UNet (ALL) and DenseNet (ALL) represent 2D and 3D model variants. The term "Other" includes all other models identified across studies (a total of 6 articles): EfficientNet, multi-linear perceptron (MLP), Deep Belief Network and long short-term memory (LSTM). In c), the Transformer, ViT and ST (ALL) include all model adaptations identified across studies: Transformer (full), Transformer encoder(s) only, Transformer blocks or layer(s), ViT (full), ViT encoder(s) only, ViT blocks, ST (full model), ST blocks, ST layers, respectively. The term "Other" includes all distinct architectural adaptations of Transf/Attention mechanisms extracted across studies: DistilGPT2 (1), channel self-attention (3), criss-cross self-attention (2), cross-attention (2), cross spatial-attention (1), multi-head self-attention (1) and spatial self-attention (1).

**Organs analysed**

We reviewed the organ under investigation across all studies (Fig. 3b). Most of the 112 studies analysed medical images from the brain (53 studies), followed by lung (20), multiple organs (20), retina (13), chest (6), neck (6), abdomen (5), heart (5), breast (3) knee (3) and pancreas (3). All other organs were examined in equal to or less than 2 studies (Fig. 3b). Studies on the top 3 most frequently analysed organs (brain, lung, multiple organs) were constantly increasing each year (Fig. 3b).

**Transformers and medical imaging**

We reviewed which Transf/Attention mechanisms were implemented per MI modality (Fig. 4a). Self-attention was mostly recruited in CT (23 studies) and MRI imaging (22), followed by ultrasound (7), retinal imaging (6), x-Rays (2) and PET-CT (2).

Transformers were the most frequent choice in MRI (16), followed by CT (4), x-Rays (3) and retinal imaging (1). ViT was mostly applied to x-Rays and retinal imaging (3 studies each), followed by CT (2). Channel- and spatial-attention mechanisms were used in MRI, CT, and x-Rays (2 studies each). ST was only used in MRI (4 studies).

**CNN and Transf/Attention combinations**

In Fig. 4b, we show that the incorporation of self-attention mechanisms was the dominant choice distributed across all CNN model types. Transformers were the second most frequent type and was used across all CNN model types, apart from VGG. ST was the third most common type and was only used in conjunction with standard CNN and UNet structures. Based on our findings, mainly "light" (simplified) Transformer blocks, encoders or layers were used across studies (Table I). Novel transfer learning strategies, multi-centre data and/ or increasingly available supercomputer facilities may encourage the use of full Transformer architectures in future work [140, 141]. However, the current hybrid models have showed performance breakthroughs across studies, highlighting them as powerful and relatively simplified (against large pre-trained models) techniques on the MIA tasks reviewed.

For standard CNN and UNet structures, all Transf/Attention mechanisms were used, except for standard attention mechanisms and ViT (Fig. 4b). For GAN models, only self-attention mechanisms were implemented. These results demonstrate that there was a large degree of variability in terms of CNN-Transf/Attention combinations across studies. Moreover, large variability was observed in the data augmentations, loss functions and metrics used to evaluate findings (Table II).

**TABLE I**

All articles grouped based on the clinical application (organ), MI modality, CNN and Transf/Attention model. To keep the information concise, details are prioritised for the top 4 organ areas (brain, lung, multiple organs and retina) in terms of prevalence, the top 2 MI modalities present per organ and the top 3 CNN (ALL) model types per organ. Transf/Attention modules were categorised to: a) Self-Attention, b) Transformer, ViT or ST (full models) and c) Transformer, ViT or ST Encoder(s), Block(s) or Layer(s). All other organs, MI modalities, CNN and Transf/Attention modules are grouped under the term "Other". Studies occurring in >1 Table cell correspond to model combinations. Missing rows of CNN models corresponds to absence of this technique per MI modality. MI: medical imaging, ViT: Vision Transformer, ST: Swin Transformer.

| Organ | MI | CNN model | Self-Attention | Transformer, ViT, ST (full) | Transformer, ViT, ST: Encoder(s), Block(s) or Layer(s) | Other Transf/ Attention |
|---|---|---|---|---|---|---|
| Brain | MRI | CNN | [36,60,77,83,98,110,115,137] | [58] | [44,46,53,54,56, 65,66,82,103] | [64] |
| | | UNet | [123] | NA | [61,65,88] | [59, 62] |
| | | GAN | [16,20,79,107,114,115,129] | NA | NA | NA |
| | | Other CNN | [15,21,38,63,94] | [39,47] | [65,70,75] | [59,97] |
| | CT | CNN | NA | NA | [65,66] | [78] |
| | | UNet | [120,144,136,138] | NA | [65] | NA |
| | | GAN | [16,79,107,136,138] | NA | NA | NA |
| | | Other CNN | [107,134] | [41] | NA | NA |
| | Other MI | UNet | [136] | NA | [88] | NA |
| | | GAN | [20,118,136] | NA | NA | NA |
| | | Other CNN | [119] | NA | NA | NA |
| Lung | CT | CNN | [18, 57, 91,99] | NA | [65] | NA |
| | | UNet | [18,95,121,136] | NA | [65] | NA |
| | | GAN | [96,114] | NA | NA | NA |
| | | Other CNN | [71,95] | NA | [43,55,85] | [84] |
| | x-Rays | CNN | NA | NA | NA | [80] |
| | | Other CNN | [92] | [42,49,90] | [43,52,55,93] | NA |
| | Other MI | CNN | NA | NA | [65] | NA |
| | | UNet | [136] | NA | [65] | NA |
| | | GAN | [118,136] | NA | NA | NA |
| Multiple organs | MRI | CNN | NA | NA | [65,66] | NA |
| | | UNet | [76] | NA | [65] | NA |
| | | Other CNN | [15,111] | NA | [65,70,109] | NA |
| | CT | CNN | [113,124] | NA | [65,66] | NA |
| | | UNet | [76,136] | NA | [65] | [127] |
| | | GAN | [124,136] | NA | NA | NA |
| | | Other CNN | [71] | [41] | [65] | [71] |
| | Other MI | CNN | [35,117] | NA | NA | NA |
| | | UNet | [108] | NA | NA | NA |
| | | GAN | [118] | NA | NA | NA |
| Retina | All Retinal imaging | CNN | [100,106,133,135] | NA | NA | [105] |
| | | UNet | [45,125,130] | NA | [51] | NA |
| | | Other CNN | [112] | [41,50] | [40,89] | NA |
| Other organ | All MI | CNN | [13,35,68,100,113,117,124, 139] | NA | [37,65,66,69] | NA |
| | | UNet | [74,76,86,108,126,128,131, 136] | NA | [65,67] | [72,102,127] |
| | | GAN | [14,19,87,118,124,132,136] | NA | NA | [19] |
| | | Other CNN | [13,15,48,71,73,96,101, 111,116,122] | [41] | [65,70,81,109] | [13,71,111, 122] |

### Downstream tasks and clinical applications

Fig. 4c illustrates all Transf/Attention components used across each downstream task. Self-attention was implemented across all organ areas (Fig. 4d). Transformer architectures were used in the brain, lung, multiple organs, heart, retina, neck, and pancreas. ViT and ST were mainly applied to a relatively limited clinical application space: lung and retina, and brain and heart, respectively. Similarly, channel- and spatial attention have been used in multiple organs, brain, lung and chest.

## C. Performance and generalization opportunities

Most proposed hybrid models have outperformed baseline and previous SOTA comparison methods, across downstream tasks. Although the evaluation metrics used differed considerably across image analysis tasks and studies making direct comparisons challenging (Table 2), there was a clear performance improvement when Transf/Attention mechanisms were used across studies. Some of the studies demonstrated either large ($\geq 5\%$) differences against the best baseline models [21, 35, 42, 46, 49, 78, 79, 94, 101, 108, 117, 121, 122, 126, 127, 135], or moderate (<5%) but consistent improvements across different metrics evaluated [13, 18, 39, 53, 54, 56, 57, 62, 70, 91, 105] and/ or data used [98, 100, 103, 105, 108].

In the following paragraphs, we detail studies that followed our 2 objective generalisation criteria (see Methods): whether a hybrid models was a) trained and/ or evaluated on large unseen testing data (>2,000 images, Table 1), b) analysed data from heterogeneous modalities, and/ or multiple modalities and/ or multiple organ areas.

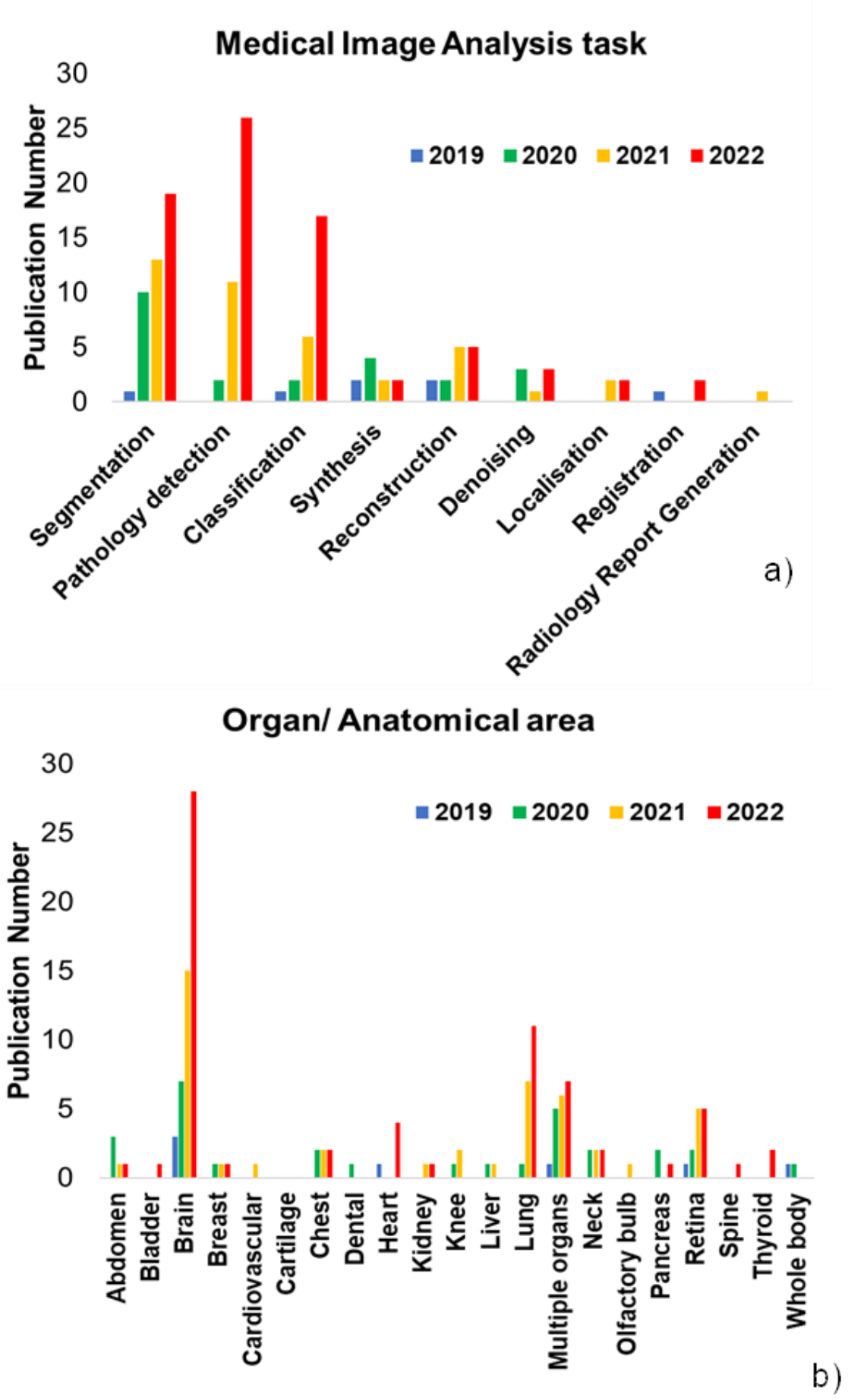

**Fig. 3**. Publication record over time for the medical image analysis task (a), and the organ/ anatomical area (clinical application) under investigation (b), respectively. The term "whole body" (Table I, Fig. 3b) by Xue et al. [118] and Dong et al [136] describes simultaneous imaging covering the entire body using a single PET-CT session, a dedicated full body technique that has been recently developed [2]. The term "multiple organs" describes all studies that imaged more than one organ in the same imaging setting [15, 19, 35, 41, 65, 66, 76, 70, 71, 76, 100, 108, 109, 111, 113, 117, 124, 127], including whole body studies [118, 136].

### Segmentation

Image segmentation is an important aspect in the field MIA, as it is a necessary intermediate step towards extracting a region of interest within the organ under investigation [142-146]. Although UNet models revolutionised medical image segmentation [147, 148], image segmentation remains an open challenge as it relies on strong supervision, hence, a large fraction of labelled data are required. However, there is a considerable "data challenge" barrier, as labels are commonly limited for MI data [2, 146]. To address this, several approaches have been proposed, such as disentangled representations for semi-supervised learning which can generate accurate segmentations by only using a small fraction of labelled data [146], or GAN techniques to obtain accurate paired synthesis of images and segmentation masks [149].

Cheng et al. proposed a multi-task methodology for simultaneous glioma segmentation from MRI images and parallel classification of genetic profiles for neuro-oncology patients [53]. They developed a CNN model with serial ResNet blocks in the encoder and decoder. Between the encoder and decoder, 2 Transformer layers were engineered. Unlike most of the MRI and CNN studies, the authors used multi-parametric MRI data for image segmentation (4 different MRI sequences). The authors compared their method against 10 baseline CNN models and demonstrated superior performance for both tasks. In the context of small but heterogeneous data analyses, Wang et al. designed a CNN encoder-decoder model with residual connections and self-attention modules connected with CNN layers in the encoder [57]. The authors demonstrate that their method outperformed all baseline models in identifying COVID-19 lung abnormalities from CT images. They also developed a zero-shot learning strategy based on the same hybrid model, in which a UNet model was applied to predict pseudo-labels in a non-labelled CT dataset, which in turn guided semi-supervised learning.

Next to limited labelled data, another challenge in medical image segmentation is the analysis of "less anatomical" and more "biophysical" imaging data, in which imaging physics are modified so that anatomical information at the pixel level is "sacrificed", to "emphasize" perfusion, functional, temporal or other biophysical information [2, 150-153]. Most segmentation algorithms are focusing on imaging sequences that contain enough anatomic (to efficiently guide semantic) representations during training [2, 153]. Shi et al. developed a powerful method that is capable to analyse 4 different parametric perfusion maps: a) cerebral blood volume, b) cerebral blood flow, c) time to maximum peak and d) mean transit time (of contrast enhancement). They developed two parallel subnetworks to analyse blood flow (a, b) and time (c, d) parameters, simultaneously. Each subnetwork included a CNN model with skip connections between the encoder and decoder. A cross-attention module was incorporated between the encoder and decoder for feature fusion. The model was compared against baseline methods (achieving higher and comparable performance, depending on the metric) and evaluated on both public and in-house data. Their method can be promising for other types of perfusion imaging data (MRI, PET, ultrasound) from which similar perfusion maps can be extracted and jointly analysed.

Rajamani et al [18] engineered a deformable attention module into a UNet model. Their model (called "DDANet") was trained and tested on a large publicly available CT COVID-19 dataset, achieving superior performance for lung infection segmentation compared to baseline models. The authors suggest that their model can be adapted to generalise in detecting small and irregular lesions for other disease areas. On a retinal image analysis study, Mou et al. developed a versatile curvilinear structure segmentation network, based on dual self-attention modules which can address both 2D and 3D retinal imaging data. In their model named as "CS2Net", they devised two channel- and spatial self-attention mechanisms to generate attention-aware features and capture long-range contextual information. By performing extensive experiments on 9 (2D and 3D) datasets, they demonstrated SOTA performances in detecting curvilinear structures from different imaging modalities. They showed that their technique can work as a generalized approach for retinal morphology analyses. Of note, such hybrid models can be impactful, since retinal imaging is not only used to assess ophthalmic pathologies, but also changes in retinal morphologies that may occur early in a broad spectrum of diseases, such as Alzheimer's [154], cardiac pathologies [155], cerebral small vessel disease [156] and others.

Xu et al. replaced 2 layers in the encoder and 1 layer in the decoder of a UNet model with self-attention mechanisms [108]. Their hybrid model achieved SOTA performance in segmenting several fetal anatomies, when compared to 6 other models. Segmenting fetal structures from ultrasound is particularly challenging due to moving and fuzzy anatomical organ boundaries [157]. Sinha et al. developed a generalizable hybrid model for segmentation of numerous abdominal, cardiovascular and brain structures by analysing different MRI sequences [111]. The authors used a ResNet model for initial feature extractions which were then fed into a stack of spatial and channel self-attention mechanisms. They demonstrated superior performance against 6 previous SOTA baselines. The model was capable to perceive a broad spectrum of anatomical (different organs) and semantic (different MRI sequences) information and can therefore potentially benefit future single- and multi-centre analyses [2, 29].

Xie et al developed a 3D UNet architecture which consisted of 2 cascading UNets both enhanced with self-attention [121]. The overall model was trained on a chronic obstructive pulmonary disease (COPD) CT dataset. Following training, the hybrid model was evaluated on COPD data and on an unseen COVID-19 dataset. The model outperformed previous techniques in the detecting several lung nodules in COPD and COVID-19 data. Following further validation using CT data from other organs, this hybrid approach can have broad applicability in terms of detecting small and irregular lesions across different diseases and organ areas.

**Pathology detection**

Pathology detection is eventually the end goal of any MIA task, with segmentation, localisation and classification commonly being designed as parallel joint tasks. In their noteworthy study, Zhou et al. proposed a cross-supervised method called REFERS, which generates image x-Ray labels from radiology reports, to perform lung pathology detection through image classification [52]. The authors employed ViT blocks composed of multi-head self-attention mechanisms, to learn joint representations from multiple radiograph views and corresponding radiology reports. Subsequently, the model performs feature fusion and employs two additional subnetworks for bidirectional visual to textual feature mapping. REFERS was first pre-trained on a source domain x-Ray dataset and then fine-tuned on 4 well-established datasets (target domain with text labels). In their transfer learning strategy, the authors performed fully supervised learning by using structured radiology report labels. Differing from other models, their technique did not require labels during pre-training. The authors also showed that their model outperformed powerful baseline models on all datasets under extremely limited supervision (1% labelled images during fine-tuning). Their model was consistently accurate in detecting several lung pathologies thus, having tremendous potential for real-world applications where labelling is substantially limited.

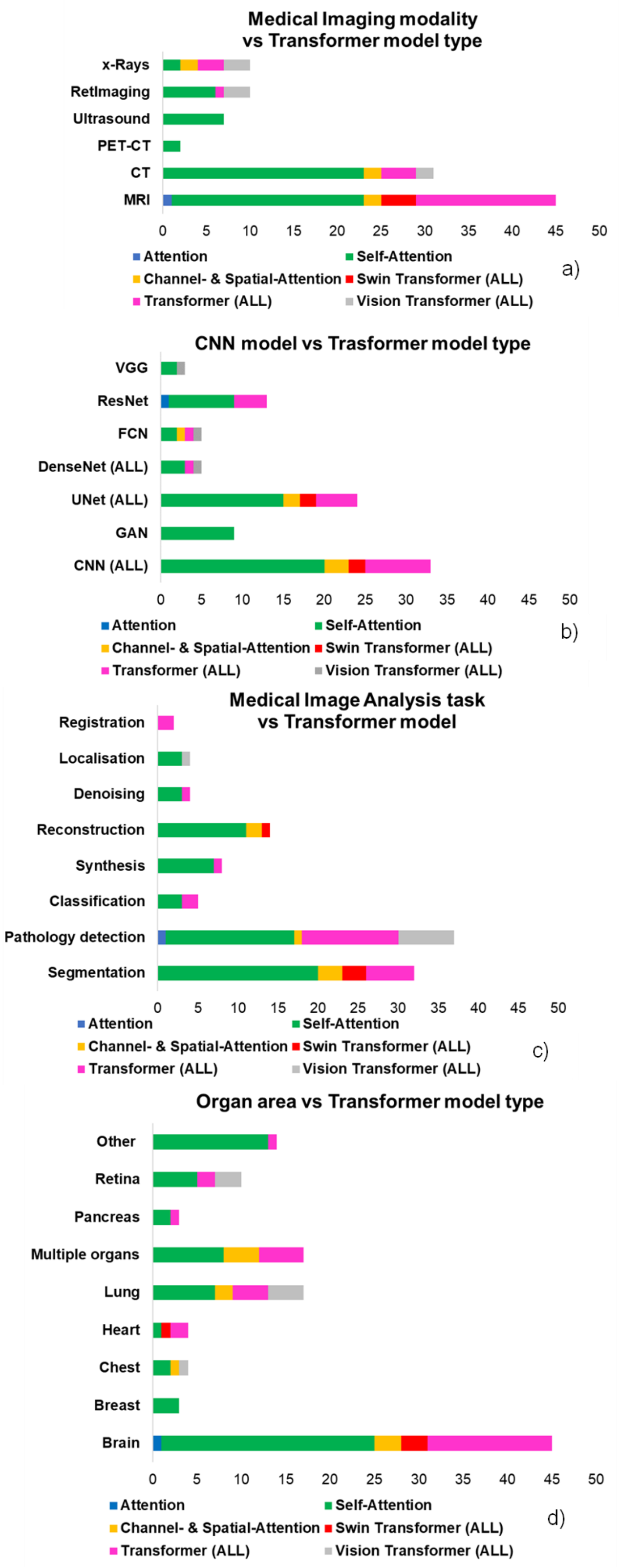

**Fig. 4.** Publication record showing combinations between the Transformer model/ component type and a) the Medical Imaging modality, b) the CNN model type, c) the Medical Image Analysis task and d) the Organ area.

**TABLE II**

All the articles grouped based on the downstream MIA task, the number of data augmentation techniques, optimiser, loss function and metric used to evaluate results. To keep the information concise, details are prioritised for the top 2 downstream tasks. All other MIA tasks are organised under the "Other MIA tasks". MIA: medical image analysis, NR: not reported, SGD: stochastic gradient descent, ACC: accuracy.

| Organ | MI | Implementation | Studies |
|---|---|---|---|
| Segmentation | Data augmentation | ≤ 3 Combined | [18,58,59,65,70,71,74,76,78,100,106, 110,119,120,125-127,131] |
| | | > 3 Combined | [13,51,53,61,67,72,105,113,116,119, 122,127] |
| | | NR | [45,57,60,62,69,98,102,103,107,108, 111,121,123,130,134] |
| | Optimiser | Adam | [13,18,45,51,53,58-62,65,70-72,74, 78,98,100,105,107,108,110, 111, 113,119,122,126, 127,131,134] |
| | | SGD | [67,76,102,106,121,123,125] |
| | | Other | [57,69,103,116,120,130] |
| | Loss function | Cross entropy-based | [18,45,51,53,57,74,103,105,106,108, 110,111,113,116,120,122,126] |
| | | Combination/Other | [13,59,60-62,65,67,69,72,76,98,100, 107,119,121,123,125,131,134] |
| | Metric | Dice | [45,51,60,67,72,98,102,103,127,131] |
| | | Combination | [13,18,53,57-59,61,62,65,69-71,74, 76,78,106-108,110,111,113, 119-123,125,126,134] |
| Pathology detection | Data augmentation | ≤ 3 Combined | [35,41-43,84,85,89,96,125,133] |
| | | > 3 Combined | [52,55,93,116] |
| | | NR | [21,36-40,44-50,54,57,63,86,87,91, 92,94,95,97] |
| | Optimiser | Adam | [21,35,37-39,41,43,45,46, 47,49,54,63,65,84,87,89,91, 92,95,97,133] |
| | | SGD | [36,42,44,48,52,86,90,125] |
| | | Other | [40,93] |
| | Loss function | Cross entropy-based | [21,35-40,42-46,52,55,57,63,84-86, 89,92-95,116,133] |
| | | Combination/Other | [41,47,48,54,55,65,87,93,125] |
| | Metric | ROC/ AUC/ACC | [21,35-43,46,48,49,52,55,63,84-87, 89-97,125,133] |
| | | Combination | [44,45,47,50-54,55,57,63,65,84,89, 91,125] |
| Other MIA tasks | Data augmentation | ≤ 3 Combined | [13,24,18,55,56,69,70,73,75,76,81,83, 85,90,92,98,100,102,103,105,109-112, 119,126,127,135,139] |
| | | > 3 Combined | [13,129] |
| | | NR | [15,16,19,20,64,66,68,77,79,80,82, 99,101,104,114,115,117,124,128,132, 136-138] |
| | Optimiser | Adam | [13,18,19,14,15,16,20,55,54,64,66,69, 70,73,75-77,79,81-83,85,88,90, 92,98-105,109,112,114, 115,117-119,124,126-129,135,137-139] |
| | | SGD | [109,135] |
| | | Other | [80,132,136] |
| | Loss function | Cross entropy-based | [20,73,81,82,135] |
| | | Combination/Other | [13-16,18,19,55,56,64,66,68-70,75-77,79,80,82,83,85,88, 90,92,98-105,109,112,114,115, 117,118,119,124,126-129,132,136-139] |
| | Metric | 1 metric (e.g., Dice, ROC, PSNR) | [13,18,19,70,73,75,77,85,98-102,104,105,109,112,117,135] |
| | | Combination | [14-16,20,56,66,68,70,79-83,88,105, 114,115,117,118,124,127-129,132, 136-139] |

To address large-scale analysis from different domains, Wood et al. developed a DenseNet-based supervised learning framework for detecting clinically relevant abnormalities from clinical T2-weighted and diffusion-weighted head MRI scans [39]. The DenseNet model was trained using a Transformer-based neuroradiology report classifier to generate a labelled dataset of 70,206 examinations from 2 UK hospitals. The Transformer model was trained using a small dataset (N= 5,000) of neuroradiology reports. The authors showed accurate, fast and generalisable classification of abnormal against normal brain MRI between hospitals. This work demonstrated the merit of CNN and Transformer synergy when combined under the same MIA pipeline.

In the context of retinal image analysis, Wang et al. have proposed the MsTGANet, a UNet model enhanced with a Transformer block that consists of a series of multi-head self-attention mechanisms incorporated in the encoder, to capture both local and global pixel interactions early in the learning process [45]. A series of channel- and spatial attention modules were also inserted between different positions of the encoder and decoder, to efficiently fuse feature semantics during training. At inference, the model predicted labels in non-labelled data, which were then used as pseudo-labels to augment the dataset, as a semi-supervised learning strategy (in which pseudo-labels were then used to guide semi-supervision). The model outperformed previous SOTA methods in supervised and semi-supervised segmentation tasks. Given the improved performance during semi-supervision, the model can potentially be used to analyse further retinal images and ophthalmic pathologies.

**TABLE III**

All the articles grouped based on the downstream MIA task, data set (public, private, both) and the data size. To keep the information concise, details are prioritised for the top 2 downstream tasks. All other MIA tasks are organised under the "Other MIA tasks". MIA: medical image analysis, NR: not reported.

| Organ | MI | Implementation | Studies |
|---|---|---|---|
| Segmentation | Data | Public | [13,18,45,51,53,57-62,65,67,70-72, 76,78,98,100,102,103,105-108, 110,111,103,121,123,125-127] |
| | | Private | [13,69,70,74,98,100,105,110,113, 116,119-122,127,130,131,134] |
| | | Both | [13,70,98,100,105,110,113,121,127] |
| | Data size | ≤ 2,000 images | [18,51,53,69,72,74,78,102,103, 106,107,110,119,120,131] |
| | | > 2,000 images and ≤ 10,000 images | [13,40,76,105,108,111,113,121,134] |
| | | > 10,000 images | [58,61,62,71,98,100] |
| | | NR | [59,60,67,70,130] |
| Pathology detection | Data | Public | [21,35,37,38,41-43,45-48,52,54,55, 57,63,65,84,85,89,90-94, 97,125,133] |
| | | Private | [35,36,39,40,44,49,50,86,87,95,96, 116] |
| | | Both | [35] |
| | Data size | ≤ 2,000 images | [21,35,50,84,85,87,125,133] |
| | | > 2,000 images and ≤ 10,000 images | [40,43,44,45,47,49,57,65,116] |
| | | > 10,000 images | [39,41,42,52,55,86,79,90] |
| | | NR | [36,37,38,46,48,54,63,91,94-97] |
| Other MIA tasks | Data | Public | [13-15,18,19,55,56,64,66,68-70,75-77,80-82,85,90,92, 98-100,102-105,109,112,114, 115,117,119,124,126,127,129,135] |
| | | Private | [13,24,26,18-20,36,55,64,68-70, 73,76,79,82,83,85,88,90, 92,98-105,127-119,126-128, 132,136-139] |
| | | Both | [13,14,18,19,55,64,68-70,76,82, 85,90,92,98-100,102-105, 117,119,126,127] |
| | Data size | ≤ 2,000 images | [15,16,20,73,75,77,83,109,112, 117,128] |
| | | > 2,000 images and ≤ 10,000 images | [14,66,79,82,88,104,125,135] |
| | | > 10,000 images | [56,80,81,101,114,118,124,129, 139] |
| | | NR | [19,74,68,132,136-138] |

Zhang et al. devised a 3D ResNet block that operated as initial feature extractor before feeding feature information into a self-attention block [21]. The authors performed several classification experiments for identifying Alzheimer's disease and mild cognitive impairment from MRI data, showing superior performance against baseline methods. Despite they focused on using T1-weighted data (mainly used for anatomical imaging and does not contain "functional" [158] or "perfusion" [159, 160] tissue information), they analysed data from both 1.5T and 3T MRI scanners, which is known that they have differences in the signal-to-noise ratio, imaging content and artefacts [1, 2, 158-160]. Since their technique was assessed on public data (ADNI), for 2 different brain pathologies and analysed data from different field MR scanners, it can potentially be promising to be assessed across further MRI data, organ areas and pathologies.

Another study by Let et al. [35], proposed a CNN encoder-decoder network connected with a self-attention mechanism (called PreSANet) to detect cancer recurrence, distant metastasis and overall patient survival for head and neck cancer patients. The model was trained on public data and was validated on various unseen datasets demonstrating good (~70%) generalisability. Chen et al. developed a ResNet model enhanced with a U-Transformer with multi-level skip-connections and outperformed SOTA baselines on anomaly detection (for pathology detection) from large MRI, CT and retinal imaging data [41]. Mondal et al. pre-trained a ViT encoder connected with a FCN layer, to discriminate COVID-19 positive cases from other pneumonia types and normal controls [55]. The model was trained on the ImageNet dataset, fine-tuned on a large collection of chest x-Ray and tested on both CT and x-Ray lung data.

Zhao et al. proposed a UNet model with residual blocks enhanced with self-attention, to classify malignant from benign thyroid nodules from ultrasound images. The model was evaluated on a large-scale dataset via extensive experiments and achieved high performance (89%) [86]. Wu et al. developed a ViT encoder and performed accurate diabetic retinopathy grading from retinal images using a large Kaggle dataset [89]. Duong et al. developed an Efficient backbone model connected with a full ViT and demonstrated accurate and generalisable detection of tuberculosis from heterogeneous x-Ray public sources [90]. Lin et al. developed a deformable ResNet model with self-attention incorporated to detect irregular and diffused lung nodules due to COVID-19 infection and showed SOTA performance in large and diverse public datasets [92]. Shome et al. developed a Transformer encoder connected with an MLP block to perform multi-classification of COVID-19 infection against other pneumonia types and nornal lung, from large x-Ray datasets [93]. Other hybrid model studies demonstrated innovative architectures and high diagnostic accuracies in the setting of pathology detection, however, using smaller datasets [47, 49, 57, 84].

**Reconstruction**

Medical image reconstruction aims to form an image representation from raw signals acquired by the scanner [2]. Reconstruction of fast acquisitions (of periodically moving organs such as the heart) and/ or low doses (e.g., CT), has important clinical applications. Using relatively small but highly diverse data, Zhou et al. developed a CNN-based method enhanced with self-attention for ultrasound image reconstruction of various organs and tissues [117]. Another study demonstrated accurate brain reconstruction by using a CNN with Transformer layers on large MRI data (>30,000 MR images) [56]. Tan et al. devised a CNN model with residual connections in which channel- and spatial-attention modules were engineered to reconstruct x-Ray images of the lung, from a large dataset (>55,000 images) [80]. Other studies focused on MRI reconstruction and demonstrated accurate and generalisable hybrid models by analysing large and diverse imaging data [114, 129, 139].

**Synthesis**

Image synthesis is an important field as it can address the need of data augmentation across different modalities [2, 161]. Yang et al. developed a CycleGAN with self-attentions for unsupervised MR-to-CT synthesis, outperforming 2 plain CycleGAN baselines [16]. In the field of MR-to-CT synthesis, Dalmaz et al. developed a series of residual Transformer blocks between the encoder and decoder of a CNN [66] and Tomar et al. developed a GAN model with ResNet blocks and self-attention modules for cardiac and brain image synthesis [107]. Wei et al. developed a first-of-its kind GAN model with self-attention in the generator and discriminator, that was able to synthesise PET-derived myelin content through the analysis of multi-sequence MRI data [20].

**Denoising**

Denoising is an important step prior to image quantification as it can enhance signal-to-noise-ratio and remove artefacts [2, 26-29]. Li et al. combined a 3D CNN model with self-attention blocks and an autoencoder perceptual loss (used as a self-supervised learning module) with CNN-based and GAN-based models. They achieved improved denoising performance against baseline models for chest and abdominal CT images [124]. Huang et al. proposed an end-to-end CycleGAN model with criss-cross self-attention and channel-attention mechanisms to reduce noise, remove artefacts and preserve anatomical structures in low-dose dental and abdominal CT images [19]. Following further validation, this can be a valuable method for future applications across multiple organs and/ or modalities. To denoise low-count PET images, Xue et al. developed a 3D GAN model with self-attention, achieving improved performance against baseline methods [118]. Their method was evaluated on large-scale PET data and showed that it can improve PET image quality, reduce motion artefacts and provide accurate diagnostic information.

**Localisation**

Image localisation focuses on detecting the location of an area of interest within MI data [26, 27]. Tao et al. proposed a ResNet model for initial feature extraction followed by a series of self-attention and cross-attention mechanisms for vertebrae CT localisation and segmentation [13]. They demonstrated accurate and generalisable performance across 2 CT datasets. Li et al. developed a DenseNet model parallelised with a ViT block to extract local and global pixel dependencies which were fused before fed into a CNN model. Their technique outperformed baseline models on classification and localisation of several lung abnormalities when trained and tested in a large x-Ray dataset (of >112,000 images) [81]. Xie et al. used a pre-trained VGG model enhanced with self-attention to enhance feature extraction before feeding this information into 2 subsequent CNN models [112]. They demonstrated accurate fovea localisation in 2 different retinal imaging datasets.

**Registration**

Image registration is the process of aligning the spatial coordinates of different images into a common geometrical coordinate system. Image registration has wide applications in multi-modal and longitudinal MIA [162, 163]. Yang et al. developed a plain Transformer encoder with an attention-based decoder model for brain MRI registration, demonstrating accurate results against baselines across 3 different datasets [75]. Song et al. proposed a CNN model with Transformer blocks consisted of modified multi-head self-attention for brain MRI registration, producing state-of-the-art registration performance [77]. Although analysing brain images from different MRI sequences is challenging, the brain is a static organ that is less prone to misregistration across modalities. Further work is required to expand towards organ areas that are subject to periodic (e.g., heart) and non-periodic (e.g., abdomen) motion, and to register images from different modalities.

**Discussion**

**A. Current opportunities and challenges**

We studied all the articles from the perspective of 4 professionals (co-authors GP, ND, CW, GY) with extensive experience in deep learning and MI. We identify general challenges and opportunities, from the multi-disciplinary perspective of developers and end-users of these hybrid models in MI. To the best of our knowledge, there is no previous review focusing on these topics and given the heterogeneous architectures of these models, more extensive studies are required in the future to develop data-driven generalization best practices for both developers and end-users. The following points can therefore guide future work and systematic reviews towards solidifying these hybrid models in further, larger and multi-centre studies.

Challenges (with mitigations, where applicable): 1) We highlighted studies that have the potential to work as generalisation frameworks. However, additional validation is required to transfer a method from one organ and/ or imaging modality to another, due to data content differences. 2) Model architectures varied considerably when similar hybrid models were compared. For example, in studies for which a UNet with self-attention were developed, there were large disparities in terms of how these individual components were combined. 3) The previous point indicates that a trial-and-error logic is currently followed for model development, based on which architecture performs optimally for a given dataset. Nevertheless, this is in the opposite direction from developing generalised models and best practices. It is important for the community to initiate discussions about the development of generalisation frameworks, based on certain data-driven boundary conditions: e.g., UNet-full Transformer for cardiac segmentation would be an optimal design if a particular data size, data content (e.g. T1-weighted MRI) and in-house computational capabilities are satisfied. Thus, solid domain generalization strategies to methodically address "why" and "how" to develop model X for data Y are required. 4) Developing harmonised implementation protocols is particularly challenging. Implementation aspects such as data augmentation, optimisers, loss functions and pre-processing differed substantially even between studies working on the same problem (e.g., CT for lung segmentation). 5) It will be challenging to develop robust interpretation mechanisms for complex local-global pattern recognition models that are not solely based on visualization maps. 6) There is an increasing trend in terms of developing causal logic in novel deep learning models, a field known as “Causal Representation Learning, CRL” [164]. The aim of CRL is to address open problems in the field such as model generalisation and transfer learning [164, 165]. Central to CRL is the discovery of high-level causal variables (objects in an image) from low-level observations (embeddings) [164]. One of the main challenges that must be addressed is how to factorise causal structures from deep learning embeddings [164, 165]. CNN-Transf/Attention models have an additional level of complexity due to learning embeddings from both local and global interactions. Thus, there must be a careful consideration regarding how to combine CNN-Transf/Attention models with causality and benefit from the advances of each other [164].

Opportunities: 1) Based on performance gains achieved, hybrid model studies can give emphasis on studying generalisation perspectives and standardisation protocols for multi-centre large-scale analyses. 2) Given diagnostic performance improvements across diverse studies, there is a potential to enhance early diagnosis and preventative medicine. 3) As of 2022, cardiovascular diseases, cancer, stroke, COVID-19, chronic respiratory diseases, diabetes, neurological diseases are the leading causes of death in the USA [166]. Most studies (>90%) in our review focused on at least 1 of the organ/ pathology areas corresponding to these leading causes, showing the potential to improve diagnosis and patient outcomes. 4) Technical versatility on multi-modal analyses can be achieved through CNN-Transf/Attention (images, natural language, molecular profiles, clinical history), which can yield useful complementary information. 5) From a clinical perspective, multi-modal data analysis can enhance precision medicine by combining patient-level information from different modalities. 6) Redirect healthcare funds towards improving treatment design and optimisation, based on multi-modal patient characteristics. 7) Accelerate the pace to introduce regulations and processes, to establish that AI for MIA can be generalisable and reproducible for certain MI and clinical applications. 8) Focus on integrating CNN-Transf/Attention with CRL to enhance model generalisability and trustworthiness in the clinical domain. 9) Develop robust transfer learning methods to fully explore the benefits of CNN-Transf/Attention models on out-of-distribution datasets.

**Importance and drawbacks**

The combination of local and global receptive fields together with reasonable computational power requirements highlights the development of CNN-Transf/Attention models as an important research direction in MIA. The large diversity of architectures even across the same downstream tasks or applications, means that for some of these methods, limited scalability may be one of the main drawbacks [52]. Furthermore, full Transformer architectures were limited in our reviewed work, mainly due to relatively small data analysed in some studies, limited computational power and/ or lack of solid transfer learning approaches for pixel-level predictions [52, 55, 141, 167]. Further work is required in the field of transfer learning techniques for model generalisation on out-of-distribution data, to utilise the benefits of full Transformer-based hybrids.

### B. Future perspectives of the post-hybrid model era

**Full transformers, ChatGPT and beyond**

The recent developments of ChatGPT large language models (LLM) induced a phenomenal disruption in the field of data analysis and AI. To date, the latest ChatGPT version is based on the GPT-4 (launched on March 2023), reported as the largest LLM trained (>170 trillion parameters) [168-170]. The main strength of GPT-4 model is that it has been trained on a diverse and broad (in terms of topics) set of internet text including books, articles and websites, using reinforcement learning from human feedback that either rewards or “punishes” the model [170]. One of its main capabilities, is that it can perform data predictions through conversational tasks (“responses” to user “queries”). ChatGPT models perform Transformer-based and self-supervised learning-derived predictions [170].

There have been some first promising approaches involving GPT models for MIA, although mainly limited to image-to-text mapping [104, 171, 172]. Wang et al. used pre-trained CNN models to extract outputs from x-Rays of the lung and applied report generator GPT models to summarise the results and derive a diagnosis in text [171]. Another study by Chen et al. used a pre-trained GPT-2 model with a visual encoding part that involved attention, to perform accurate image captioning as evaluated on natural images and X-Ray data [172]. Jeblick et al. developed a GPT-based technique that focused on simplifying radiology reports but without using imaging data as inputs [173].

Although it can be anticipated that GPT models may expand towards MIA, there are several limitations that need to be considered. First, to the best of our knowledge, there is no GPT-based MIA yet on dense image-level predictions for the MIA tasks we have presented. Local receptive fields that are based on CNN feature extractors may be necessary to perform detailed image analyses, pointing towards the direction of heavier “hybrid models” in the future (CNN-GPT). On that note, it is unknown whether existing self-supervision modules within GPT models may be enough to predict complex organ and tissue pathologies from “high-content” data such as medical images, without the incorporation of “computer vision” CNN components. Furthermore, one important limitation of GPT models is the so called “hallucination effect”, which describes the tendency of GPT models to “invent” a term eventually giving “incorrect” responses [174]. This can be the case for domains in which GPT models have been less or not yet specialized. Due to regulatory, ethical and organisational considerations from clinical and private MI data owners, we are still at infant stages regarding multi-centre large-scale data analyses that need to be available as data sources for such open code or multi-centre fine-tuning strategies. In addition, the co-existence of available MI and text data is commonly low. To date, given the demonstrated performance of current hybrid models in our review, convolutional Transf/Attention may be “all you need” in medical imaging.

**Transfer learning coming from the future**

An important yet unsolved aspect in MIA is the democratisation of modelling techniques and data. Transfer learning strategies focusing on increasing performance and generalisability while reducing computational power [141], can serve as democratisation vehicles. However, it is known that transfer learning for image-level predictions has been limited in MIA, compared to "smaller new" models that are trained de novo.

Among a large amount of studies demonstrating new models, we highlighted articles that showed robust pre-training with wide fine-tuning on large domain datasets with SOTA performance on testing data [52, 55]. In their influential study, Liu et al. recently proposed "ConvNext" as a new pure CNN technique which involved several ST-inspired adaptations in the model design and transfer learning method [141]. Some of these ST-inspired adaptations were: same augmentation protocol, network width increase, bottleneck model inversion, kernel size enlargement, use of fewer activation functions and normalisation layers. Using ConvNext, they outperformed ST on ImageNet classification tasks while using comparable computing resources. Radford et al. adapted Transformer and ResNet/ ViT models for jointly pre-training paired text and images, respectively [167]. By training on web data of 400 million image-text pairs, they demonstrate that can learn image captions from text which can be used as labels for image classification, showcasing a scalable and efficient process to learn image representations. Following pre-training, the text model can describe new visual concepts allowing zero-shot transfer to new tasks and data. Further work on these directions will be particularly important to improve model design, pre-training (on out-of-domain data) and fine-tuning (on domain data) techniques towards efficiently democratising large hybrid models and data access.

## Conclusions

In conclusion, hybrid models led to performance gains while demonstrating a big range of generalisation opportunities based on either their large-scale, multi-modal, heterogeneous and/ or broad span of clinical applications. The main challenge of these techniques is to align their large architectural diversity with the current technical and clinical needs in precision and preventative medicine. Based on the opportunities that we have emphasised, we aim to encourage further work on data-driven generalisation frameworks, to develop criteria for the future design of these powerful hybrid techniques. We also seek to inspire further work in the field of transfer learning for generalisation on out-of-distribution data so that models and data can be further democratised.

Our review demonstrates the benefits from the co-pollination of CNN and Transformer-inspired models which can open new horizons to further exploit CNN and full Transformers and LLM. Next to these opportunities, our review demonstrated that the benefits of CNN-Transf/Attention outweigh the challenges and may therefore be "all you need" for future validation and standardisation processes in clinical imaging.

## References

1. J. Beutel, et al. Handbook of Medical Imaging. Bellingham, WA, USA: SPEI Press. 2000, 1.
2. S. Kevin Zhou, et al., A review of deep learning in medical imaging: Imaging traits, technology trends, case studies with progress highlights, and future promises. Proceedings of the IEEE. 109(5): 820-838, 2020.
3. Radiology Society of North America (RSNA). Quantitative Imaging Biomarkers Aliance (QIBA) [Online]. Available: https://www.rsna.org/research/quantitative-imaging-biomarkers-alliance.
4. NM. deSouza, et al., Validated imaging biomarkers as decision-making tools in clinical trials and routine practice: current status and recommendations from the EIBALL* subcommittee of the European Society of Radiology (ESR). Insights Imaging. 10(87), 2019.
5. European Society of Radiology (ESR). ESR Statement on the Validation of Imaging Biomarkers. Insights Imaging. 11 (76), 2020.
6. J. O'Connor, et al. Imaging biomarker roadmap for cancer studies. Nat Rev Clin Oncol. 14: 169–186, 2017.
7. A. Krizhevsky, I. Sutskever, Hinton GE. ImageNet classification with deep convolutional neural networks. in Proc. Adv. Neural Inf. Process Syst. 1097–1105, 2012.
8. D. Bahdanau, K. Cho, Y. Bengio. Neural machine translation by jointly learning to align and translate. CoRR, abs/1409.0473, 2014.
9. J. Cheng, L. Dong, M. Lapata. Long short-term memory-networks for machine reading. arXiv preprint arXiv:1601.06733, 2016.
10. A. Dosovitskiy, L. Beyer, A. Kolesnikov, D. Weissenborn, et al. An image is worth 16x16 words: Transformers for image recognition at scale. arXiv preprint: https://arxiv.org/abs/2010.11929, 2020.
11. Z. Liu, Y. Lin, Y. Cao, H. Hu, Y. Wei, Z. Zhang, S. Lin, B. Guo. Swin transformer: Hierarchical vision transformer using shifted windows. In ICCV, 2021.
12. Z. Liu, H. Hu, Y. Lin, et al. Swin transformer v2: scaling up capacity and resolution. arXiv preprint: https://arxiv.org/abs/2111.09883, 2021.
13. R. Tao, W. Liu, G. Zheng. Spine-transformers: Vertebra labeling and segmentation in arbitrary field-of-view spine CTs via 3D transformers. Medical Image Analysis. 75: 102258, 2022.
14. S. Bera, P.K. Biswas. Noise conscious training of non Local neural network powered by self attentive spectral normalized markovian patch GAN for low dose CT denoising. IEEE Trans Med Imaging. 40(12): 3663-3673, 2021.
15. T. Du, H. Zhang, Y. Li, S. Pickup, et al. Adaptive convolutional neural networks for accelerating magnetic resonance imaging via k-space data interpolation. Medical Image Analysis. 72: 102098, 2021.
16. H. Yang, et al. Unsupervised MR-to-CT Synthesis Using Structure-Constrained CycleGAN. IEEE Transactions on Medical Imaging. 39(12): 4249-4261, 2020.
17. F. Shamshad, S. Khan, S. Waqas Zamir, et al. Transformers in medical imaging: a survey. arXiv preprint: https://arxiv.org/abs/2201.09873, 2022.
18. K.T. Rajamani, H. Siebert, M.P. Heinrich. Dynamic deformable attention network (DDANet) for COVID-19 lesions semantic segmentation. Journal of Biomedical Informatics. 119, 2021.
19. Z. Huang, et al. CaGAN: A cycle-consistent generative adversarial network with attention for low-dose CT Imaging. IEEE Transactions on Computational Imaging. 6: 1203-1218, 2020.
20. W. Wei, E. Poirion, B. Bodini, M. Tonietto, S. Durrleman, O. Colliot, B. Stankoff, N. Ayache. Predicting PET-derived myelin content from multisequence MRI for individual longitudinal analysis in multiple sclerosis. NeuroImage. 223: 117308, 2020.
21. Zhang X, Han L, Zhu W, Sun L, Zhang D. An Explainable 3D Residual Self-Attention Deep Neural Network For Joint Atrophy Localization and Alzheimer's Disease Diagnosis using Structural MRI. IEEE J Biomed Health Inform. doi: 10.1109/JBHI.2021.3066832, 2021.
22. R.R. Selvaraju, M. Cogswell, A. Das, R. Vedantam, D. Parikh, D. Batra. Grad-cam: Visual explanations from deep networks via gradient-based localization. In Proceedings of the IEEE international conference on computer vision. pp: 618–626, 2017.
23. O. Dalmaz, M. Yurt, T. Cukur. ResViT: Residual vision transformers for multi-modal medical image synthesis. IEEE Transactions on Medical Imaging. 41(10):2598-2614, 2022.
24. H. Chefer, S. Gur, L. Wolf. Transformer interpretability beyond attention visualization. CVPR, 2021.
25. L. Yu, L. Maozhen, J. Changjun. Generating self-attention activation maps for visual interpretations of convolutional neural networks. Neurocomputing. 490: 206-216, 2022.
26. X. Yi, W. Walia, P. Babyn. Generative adversarial network in medical imaging: A review. Medical Image Analysis. 58: 101552, 2019.
27. V. Cheplygina, M. de Bruijne, J.P.W. Pluim. Not-so-supervised: A survey of semi-supervised, multi-instance, and transfer learning in medical image analysis. Medical Image Analysis. 54: 280–296, 2019.
28. J.S. Duncan, M.F. Insana, N. Ayache. Biomedical imaging and analysis in the age of big data and deep learning. Proc. IEEE. 108(1): 3-10, 2020.
29. G. Haskins, U. Kruger, P. Yan. Deep learning in medical image registration: A survey. Mach. Vis. Appl. 31(1), 2020.

30. K. He, C. Gan, Z. Li, et al. Transformers in medical image analysis, Intelligent Medicine. ISSN 2667-1026, 2022.
31. N. Linna, C.E. Kahn. Applications of natural language processing in radiology: A systematic review. International Journal of Medical Informatics. 163:104779, 2022.
32. X. Chen, X. Wang, K. Zhang, et al. Recent advances and clinical applications of deep learning in medical image analysis. Medical Image Analysis. 79: 102444, 2022.
33. A. Parvaiz, M.A. Khalid, R. Zafar, et al. Vision Transformers in medical computer vision-A contemplative retrospection. Engineering Applications of Artificial Intelligence.122: 106126, 2023.
34. D. Moher, A. Liberati, J. Tetzlaff, D.G. Altman. The PRISMA Group. Preferred Reporting Items for Systematic Reviews and Meta-Analyses: The PRISMA Statement. PLoS Med. 6(7):e1000097, 2009.
35. W.T. Le, E. Vorontsov, F.P. Romero, et al. Cross-institutional outcome prediction for head and Neck cancer patients using self-attention neural networks. Scientific Reports. 12(1):3183, 2022.
36. K.H. Oh, I.S. Oh, U. Tsogt, et al. Diagnosis of schizophrenia with functional connectome data: a graph-based convolutional neural network approach. BMC Neuroscience. 23(1):5, 2022.
37. Z. Zhou, L. Yu, S. Tian, et al. Local-global multiple perception based deep multi-modality learning for sub-type of esophageal cancer classification. Biomedical Signal Processing and Control. 77: 103757, 2022.
38. L. Wang, W. Yuan, L. Zeng, et al. Dementia analysis from functional connectivity network with graph neural networks. Information Processing and Management. 59(3):102901, 2022.
39. D.A. Wood, S. Kafiabadi, A.A. Busaidi, et al. Deep learning models for triaging hospital head MRI examinations. Medical Image Analysis. 78: 102391, 2022.
40. Z. Jiang, L. Wang, Q. Wu, et al. Computer-aided diagnosis of retinopathy based on vision transformer. Journal of Innovative Optical Health Sciences. 15(2):2250009, 2022.
41. L. Chen, Z. You, N. Zhang, J. Xi, X. Le. UTRAD: Anomaly detection and localization with U-Transformer. Neural Networks. 147:53-62, 2022.
42. S. Rajaraman, G. Zamzmi, L.R. Folio, S. Antani. Detecting Tuberculosis-Consistent Findings in Lateral Chest X-Rays Using an Ensemble of CNNs and Vision Transformers. Frontiers in Genetics. 13: 864724, 2022.
43. M.M. Al Rahhal, Y. Bazi, R.M. Jomaa, A. Alshibli, N. Alajlan, M.L. Mekhalfi, F. Melgani. COVID-19 Detection in CT/X-ray Imagery Using Vision Transformers. Journal of Personalized Medicine. 12(2):310, 2022.
44. J. Zhang, C. Li, G. Liu, M. Min, C. Wang, J. Li, Y. Wang, H. Yan, Z. Zuo, W. Huang, H. Chen. A CNN-transformer hybrid approach for decoding visual neural activity into text. Computer Methods and Programs in Biomedicine. 214: 106586, 2022.
45. M. Wang, W. Zhu, F. Shi, et al. MsTGANet: Automatic Drusen Segmentation from Retinal OCT Images. IEEE Transactions on Medical Imaging. 41(2):394-406, 2022.
46. G. Zheng, Y. Zhang, Z. Zhao, et al. A transformer-based multi-features fusion model for prediction of conversion in mild cognitive impairment. Methods. 204: 241-248, 2022.
47. S. He, Y. Feng, G. Ellen Grant, Y. Ou. Deep Relation Learning for Regression and Its Application to Brain Age Estimation. IEEE Transactions on Medical Imaging. 41(9): 2304-2317, 2022.
48. T.S. Chandraraju, A. Jeyaprakash. Categorization of Breast masses based on deep belief network parameters optimized using chaotic krill herd optimization algorithm for frequent diagnosis of Breast abnormalities. International Journal of Imaging Systems and Technology. 32(5):1561-1576, 2022.
49. J. Cheng, J. Sollee, C. Hsieh, et al. COVID-19 mortality prediction in the intensive care unit with deep learning based on longitudinal Chest X-rays and clinical data. European Radiology. 32:(7)4446-4456, 2022.
50. R. Otsuki, O. Sugiyama, Y. Mori, et al. Integrating Preprocessing Operations into Deep Learning Model: Case Study of Posttreatment Visual Acuity Prediction. Advanced Biomedical Engineering. 11:16-24, 2022.
51. D. Chen, W. Yang, L. Wang, et al. PCAT-UNet: UNet-like network fused convolution and transformer for Retinal vessel segmentation. PLoS ONE. 17(11):e0262689, 2022.
52. H.Y. Zhou, X. Chen, Y. Zhang, R. Luo, L. Wang, Y. Yu. Generalized radiograph representation learning via cross-supervision between images and free-text radiology reports. Nature Machine Intelligence. 4(1):32-40, 2022.
53. J. Cheng, J. Liu, H. Kuang, J. Wang. A Fully Automated Multimodal MRI-based Multi-task Learning for Glioma Segmentation and IDH Genotyping. IEEE Transactions on Medical Imaging. 41(6):1520-1532, 2022.
54. S. He, P.E. Grant, Y. Ou. Global-Local Transformer for Brain Age Estimation. IEEE Transactions on Medical Imaging. 41(1):213-224, 2022.
55. A.K. Mondal, A. Bhattacharjee, P. Singla, A.P. Prathosh. XViTCOS: Explainable Vision Transformer Based COVID-19 Screening Using Radiography. IEEE Journal of Translational Engineering in Health and Medicine. 10:1100110, 2022.
56. J. Huang, Y. Fang, Y. Wu, H. Wu, Z. Gao, Y. Li, J.D. Ser, J. Xia J, G. Yang. Swin transformer for fast MRI. Neurocomputing. 493: 281-304, 2022.
57. X. Wang, Y. Yuan, D. Guo, et al. SSA-Net: Spatial self-attention network for COVID-19 pneumonia infection segmentation with semi-supervised few-shot learning. Medical Image Analysis. 79: 102459, 2022.
58. J. Liu, J. Zheng, G. Jiao. Transition Net: 2D backbone to segment 3D Brain tumor. Biomedical Signal Processing and Control. 75: 103622, 2022.
59. L. Wu, S. Hu, C. Liu. MR Brain segmentation based on DE-ResUnet combining texture features and background knowledge. Biomedical Signal Processing and Control. 75: 103541, 2022.
60. C. Laiton-Bonadiez, G. Sanchez-Torres, J. Branch-Bedoya. Deep 3D Neural Network for Brain Structures Segmentation Using Self-Attention Modules in MRI Images. Sensors. 22(7):2559, 2022.
61. J. Liang, C. Yang, M. Zeng, X. Wang. TransConver: Transformer and convolution parallel network for developing automatic Brain tumor segmentation in MRI images. Quantitative Imaging in Medicine and Surgery. 23(5):e13597, 2022.
62. M. Sheng, W. Xu, J. Yang, Z. Chen. Cross-Attention and Deep Supervision UNet for Lesion Segmentation of Chronic Stroke. Frontiers in Neuroscience. 482: 82-97, 2022.
63. M. Jiang, B. Yan, Y. Li, J. Zhang, T. Li, W. Ke. Image Classification of Alzheimer's Disease Based on External-Attention Mechanism and Fully Convolutional Network. Brain Sciences. 12(3):319, 2022.
64. L. Wang, H. Zhu, Z. He, Y. Jia, J. Du. Adjacent slices feature transformer network for single anisotropic 3D Brain MRI image super-resolution. Biomedical Signal Processing and Control. 72:103339, 2022.
65. T. Dhamija, A. Gupta, S. Gupta, R. Anjum Katarya, G. Singh. Semantic segmentation in medical images through transfused convolution and transformer networks. Applied Intelligence. 53(1):1132-1148, 2022.
66. O. Dalmaz, M. Yurt, T. Cukur. ResViT: Residual vision transformers for multi-modal medical image synthesis. IEEE Transactions on Medical Imaging. 41(10): 2598-2614, 2022.
67. Z. Fu, J. Zhang, R. Luo, Y. Sun, D. Deng, L. Xia. TF-Unet: An automatic cardiac MRI image segmentation method. Mathematical Biosciences and Engineering. 19(5): 5207-5222, 2022.
68. H. Wang, X. Zhao, W. Liu, L.C. Li, J. Ma. L. Guo. Texture-aware dual domain mapping model for low-dose CT reconstruction. Medical Physics. 49(6):3860-3873, 2022.
69. J. Wang, S. Wang, W. Liang, N. Zhang, Y. Zhang. The auto-segmentation for cardiac structures using a dual-input deep learning network based on vision saliency and transformer. Journal of Applied Clinical Medical Physics. 23(5):e13597, 2022.
70. D. Karimi, H. Dou, A. Gholipour. Medical Image Segmentation Using Transformer Networks. IEEE Access. 10: 29322-29332, 2022.
71. M. Ma, H. Xia, Y. Tan, H. Li, S. Song. HT-Net: hierarchical context-attention transformer network for medical ct image segmentation. Applied Intelligence. 52(9):10692-10705, 2022.
72. H. Cui, L. Jiang, C. Yuwen, Y. Xia, Y. Zhang. Deep U-Net architecture with curriculum learning for myocardial pathology segmentation in multi-sequence cardiac magnetic resonance images. Knowledge-Based Systems. 249: 108942, 2022.
73. N. Cahan, E.M. Marom, S. Soffer, Y. Barash, E. Konen, E. Klang, H. Greenspan. Weakly supervised attention model for RV strain classification from volumetric CTPA scans. Computer Methods and Programs in Biomedicine. 220: 106815, 2022.
74. X. Wang, L. Wang, Y. Sheng, C. Zhu, N. Jiang, C. Bai, M. Xia, Z. Shao, Z. Gu, X. Huang, R. Zhao, Z. Liu. Automatic and accurate segmentation of peripherally inserted central catheter (PICC) from Chest X-rays using multi-stage attention-guided learning. Neurocomputing. 48: 82-97, 2022.
75. T. Yang, X. Bai, X. Cui, Y. Gong, L. Li. TransDIR: Deformable imaging registration network based on transformer to improve the feature extraction ability. Medical Physics. 49(2):952-965, 2022.
76. J. Xie, R. Zhu, Z. Wu, J. Ouyang. FFUNet: A novel feature fusion makes strong decoder for medical image segmentation. IET Signal Processing. 16(5):501-514, 2022.
77. L. Song, G. Liu, M. Ma. TD-Net:unsupervised medical image registration network based on Transformer and CNN. Applied Intelligence. 52(15):18201-18209, 2022.
78. T. Shi, H. Jiang, B. Zheng. C2MA-Net: Cross-Modal Cross-Attention Network for Acute Ischemic Stroke Lesion Segmentation Based on CT Perfusion Scans. IEEE Transactions on Biomedical Engineering. 69(1):108-118, 2022.
79. C. Wang, J. Uh, T.E. Merchant, C.H. Hua, S. Acharya. Facilitating MR-Guided Adaptive Proton Therapy in Children Using Deep Learning-Based Synthetic CT. International Journal of Particle Therapy. 8(3):11-20, 2022.
80. Z. Tan, J. Li, H. Tao, S. Li, Y. Hu. XctNet: Reconstruction network of volumetric images from a single X-ray image. Computerized Medical Imaging and Graphics. 98: 102067, 2022.
81. F. Li, L. Zhou, Y. Wang, C. Chen, S. Yang, F. Shan, L. Liu. Modeling long-range dependencies for weakly supervised disease classification and localization on Chest X-ray. Quantitative Imaging in Medicine and Surgery. 12(6):3364-3378, 2022.
82. J.X. Wang, Y. Li, X. Li, Z.H. Lu. Alzheimer's Disease Classification Through Imaging Genetic Data With IGnet. Frontiers in Neuroscience. 16: 846638, 2022.
83. J.S. Hong, I. Hermann, F.G. Zöllner, L.R. Schad, S.J. Wang, W.K. Lee, Y.L. Chen, Y. Chang, Y.T. Wu. Acceleration of Magnetic Resonance Fingerprinting

Reconstruction Using Denoising and Self-Attention Pyramidal Convolutional Neural Network. Sensors. 22(3):1260, 2022.
84. M. Al-Shabi, K. Shak, M. Tan. ProCAN: Progressive growing channel attentive non-local network for Lung nodule classification. Pattern Recognition. 122: 108309, 2022.
85. D. Liu, F. Liu, Y. Tie, L. Qi, F. Wang. Res-trans networks for Lung nodule classification. International Journal of Computer Assisted Radiology and Surgery. 17(6):1059-1068, 2022.
86. S. Zhao, Y. Chen, K. Yang, K. Yang, Y. Luo, B. Ma, Y. Li. A Local and Global Feature Disentangled Network: Toward Classification of Benign-malignant Thyroid Nodules from Ultrasound Image. IEEE Transactions on Medical Imaging. 41(6):1497-1509, 2022.
87. J. Zhao, X. Zhou, G. Shi, N. Xiao, K. Song, J. Zhao, R. Hao, K. Li. Semantic consistency generative adversarial network for cross-modality domain adaptation in ultrasound Thyroid nodule classification. Applied Intelligence. 52(9):10369-10383, 2022.
88. L. Zhang, Z. Xiao, C. Zhou, J. Yuan, Q. He, Y. Yang, X. Liu, D. Liang, H. Zheng, W. Fan, X. Zhang, Z. Hu. Spatial adaptive and transformer fusion network (STFNet) for low-count PET blind denoising with MRI. Medical Physics. 49(1):343-356, 2022.
89. J. Wu, R. Hu, Z. Xiao, J. Chen, J. Liu. Vision Transformer-based recognition of diabetic retinopathy grade. Medical Physics. 48(12):7850-7863, 2021.
90. L.T. Duong, N.H. Le, T.B. Tran, V.M. Ngo, P.T. Nguyen. Detection of tuberculosis from Chest X-ray images: Boosting the performance with vision transformer and transfer learning. Expert Systems with Applications. 184:115519, 2021.
91. Z. Al Nazi, F. Rabbi Mashrur, M.A. Islam, S. Saha. Fibro-CoSANet: Pulmonary fibrosis prognosis prediction using a convolutional self attention network. Physics in Medicine and Biology. 66(22): 225013, 2021.
92. Z. Lin, Z. He, S. Xie, X. Wang, J. Tan, J. Lu, B. Tan. AANet: Adaptive Attention Network for COVID-19 Detection from Chest X-Ray Images. IEEE Transactions on Neural Networks and Learning Systems. 32(11):4781-4792, 2021.
93. D. Shome, T. Kar, S.N. Mohanty, P. Tiwari, K. Muhammad, A. Altameem, Y. Zhang, A.K.J. Saudagar. Covid-transformer: Interpretable covid-19 detection using vision transformer for healthcare International Journal of Environmental Research and Public Health. 18(21):11086, 2021.
94. Z. Wang, D. Peng, Y. Shang, J. Gao. Autistic Spectrum Disorder Detection and Structural Biomarker Identification Using Self-Attention Model and Individual-Level Morphological Covariance Brain Networks. Frontiers in Neuroscience. 15:756868, 2021.
95. Y. Fu, P. Xue, E. Dong. Densely connected attention network for diagnosing COVID-19 based on Chest CT. Computers in Biology and Medicine. 137:104857, 2021.
96. F. Rundo, M. Bersanelli, V. Urzia, A. Friedlaender, O. Cantale, G. Calcara, A. Addeo, G.L. Banna. Three-Dimensional Deep Noninvasive Radiomics for the Prediction of Disease Control in Patients With Metastatic Urothelial Carcinoma treated With Immunotherapy. Clinical Genitourinary Cancer. 19(5):396-404, 2021.
97. H. Sun, A. Wang, W. Wang, C. Liu. An improved deep residual network prediction model for the early diagnosis of alzheimer's disease. Sensors. 21(12):4182, 2021.
98. S. Estrada, R. Lu, K. Diers, et al. Automated Olfactory bulb segmentation on high resolutional T2-weighted MRI. NeuroImage. 242:118464, 2021.
99. H. Xie, T. Zhang, W. Song, S. Wang, H. Zhu, R. Zhang, W. Zhang, Y. Yu, Y. Zhao. Super-resolution of Pneumocystis carinii pneumonia CT via self-attention GAN. Computer Methods and Programs in Biomedicine. 212: 106467, 2021.
100. Y. Li, J. Yang, J. Ni, A. Elazab, J. Wu. TA-Net: Triple attention network for medical image segmentation. Computers in Biology and Medicine. 137:104836, 2021.
101. X. Qu, G. Yan, D. Zheng, S. Fan, Q. Rao, J. Jiang. A Deep Learning-Based Automatic First-Arrival Picking Method for Ultrasound Sound-Speed Tomography. IEEE Transactions on Ultrasonics, Ferroelectrics, and Frequency Control. 68(8):2675-2686, 2021.
102. M. Kim, B.D. Lee. Automatic Lung segmentation on Chest x-rays using self-attention deep neural network. Sensors. 21(2):1-12, 2021.
103. Q. Sun, N. Fang, Z. Liu, L. Zhao, Y. Wen, H. Lin. HybridCTrm: Bridging CNN and Transformer for Multimodal Brain Image Segmentation. Journal of Healthcare Engineering. 7467261, 2021.
104. O. Alfarghaly, R. Khaled, A. Elkorany, M. Helal, A. Fahmy. Automated radiology report generation using conditioned transformers. Informatics in Medicine Unlocked. 24:100557, 2021.
105. L. Mou, Y. Zhao, H. Fu, Y. Liu, et al. CS2-Net: Deep learning segmentation of curvilinear structures in medical imaging. Medical Image Analysis. 67:101874, 2021.
106. C.Z. Wu, J. Sun, J. Wang, L.F. Xu, S. Zhan. Encoding-decoding Network With Pyramid Self-attention Module For Retinal Vessel Segmentation. International Journal of Automation and Computing. 18(6):973-980, 2021.
107. D. Tomar, M. Lortkipanidze, G. Vray, B. Bozorgtabar, J.P. Thiran. Self-Attentive Spatial Adaptive Normalization for Cross-Modality Domain Adaptation. IEEE Transactions on Medical Imaging. 40(10):2926-2938, 2021.
108. L. Xu, S. Gao, L. Shi, B. Wei, X. Liu, J. Zhang, Y. He. Exploiting vector attention and context prior for ultrasound image segmentation. Neurocomputing. 454: 461-473, 2021.
109. Y. Dai, Y. Gao, F. Liu. Transmed: Transformers advance multi-modal medical image classification. Diagnostics. 11(8):1384, 2021.
110. T. Zhong, F. Zhao, Y. Pei, et al. DIKA-Nets: Domain-invariant knowledge-guided attention networks for Brain skull stripping of early developing macaques. NeuroImage. 227:117649, 2021.
111. A. Sinha, J. Dolz. Multi-Scale Self-Guided Attention for Medical Image Segmentation. IEEE Journal of Biomedical and Health Informatics. 25(1):121-130, 2021.
112. R. Xie, J. Liu, R. Cao, C.S. Qiu, J. Duan, J. Garibaldi, G. Qiu. End-to-End Fovea Localisation in Colour Fundus Images with a Hierarchical Deep Regression Network. IEEE Transactions on Medical Imaging. 40(1):116-128, 2021.
113. Z. Zhang, T. Zhao, H. Gay, W. Zhang, B. Sun. Weaving attention U-net: A novel hybrid CNN and attention-based method for organs-at-risk segmentation in head and Neck CT images. Medical Physics. 48(11):7052-7062, 2021.
114. W. Zhou, H. Du, W. Mei, L. Fang. Spatial orthogonal attention generative adversarial network for MRI reconstruction. Medical Physics. 2021; 48(2):627-639.
115. Z. Hu, H. Liu, Z. Li, Z. Yu. Cross-Model Transformer Method for Medical Image Synthesis. Complexity. 5624909, 2021.
116. S. Lee, E. Kim, J.S. Bae, J.H. Kim, S. Yoon. Robust End-to-End Focal Liver Lesion Detection Using Unregistered Multiphase Computed Tomography Images. IEEE Transactions on Emerging Topics in Computational Intelligence. 7(2):319-327, 2021.
117. Z. Zhou, Y. Wang, Y. Guo, X. Jiang, Y. Qi. Ultrafast Plane Wave Imaging with Line-Scan-Quality Using an Ultrasound-Transfer Generative Adversarial Network. IEEE Journal of Biomedical and Health Informatics. 24(4):943-956, 2020.
118. H. Xue, Y. Teng, C. Tie, et al. A 3D attention residual encoder–decoder least-square GAN for low-count PET denoising. Nuclear Instruments and Methods in Physics Research, Section A: Accelerators, Spectrometers, Detectors and Associated Equipment. 983:164638, 2020.
119. J.M.J. Valanarasu, R. Yasarla, P. Wang, et al. Learning to Segment Brain Anatomy from 2D Ultrasound with Less Data. IEEE Journal on Selected Topics in Signal Processing. 14(6):1221-1234, 2020.
120. Z. Kuang, X. Deng, L. Yu, H. Wang, T. Li, S. Wang. Ψ-Net: Focusing on the border areas of intracerebral hemorrhage on CT images. Computer Methods and Programs in Biomedicine. 194: 105546, 2020.
121. W. Xie, C. Jacobs, J.P. Charbonnier, B. Van Ginneken. Relational Modeling for Robust and Efficient Pulmonary Lobe Segmentation in CT Scans. IEEE Transactions on Medical Imaging. 39(8):2664-2675, 2020.
122. B. Lei, S. Huang, H. Li, R. Li, et al. Self-co-attention neural network for anatomy segmentation in Whole Breast ultrasound. Medical Image Analysis. 64:101753, 2020.
123. X. Jia, Y. Liu, Z. Yang, D. Yang. Multi-modality self-attention aware deep network for 3D biomedical segmentation. BMC Medical Informatics and Decision Making. 20:119, 2020.
124. M. Li, W. Hsu, X. Xie, J. Cong, W. Gao. SACNN: Self-Attention Convoluti-onal Neural Network for Low-Dose CT Denoising with Self-Supervised Perceptual Loss Network. IEEE Transactions on Medical Imaging. 983:164638, 2020.
125. S. Li, C. Ge, X. Sui, Y. Zheng, W. Jia. Channel and spatial attention regression network for cup-to-disc ratio estimation. Electronics. 9(6):909, 2020.
126. T. Fan, G. Wang, Y. Li, H. Wang. Ma-net: A multi-scale attention network for Liver and tumor segmentation. IEEE Access. 8:179656-179665, 2020.
127. S. Gou, N. Tong, S. Qi, S. Yang, R. Chin, K. Sheng. Self-channel-and-spatial-attention neural network for automated multi-organ segmentation on head and Neck CT images. Physics in Medicine and Biology. 65 (24):245034, 2020.
128. Y. Wu, D. Li, L. Xing, G. Gold. Deriving new soft tissue contrasts from conventional MR images using deep learning. Magnetic Resonance Imaging. 74:121-127, 2020.
129. Z. Yuan, M. Jiang, Y. Wang, et al. SARA-GAN: Self-Attention and Relative Average Discriminator Based Generative Adversarial Networks for Fast Compressed Sensing MRI Reconstruction. Frontiers in Neuroinformatics. 14:611666, 2020.
130. P. Zhong, J. Wang, Y. Guo, X. Fu, R. Wang. Multiclass Retinal disease classification and lesion segmentation in OCT B-scan images using cascaded convolutional networks. Applied Optics. 59(33):10312-10320, 2020.
131. Y. Liu, Y. Lei, Y. Fu, et al. CT-based multi-organ segmentation using a 3D self-attention U-net network for pancreatic radiotherapy. Medical Physics. 47(9):4316-4324, 2020.
132. Y. Liu, Y. Lei, T. Wang, et al. CBCT-based synthetic CT generation using deep-attention cycleGAN for pancreatic adaptive radiotherapy. Medical Physics. 47(6):2472-2483, 2020.
133. V. Das, E. Prabhakararao, S. Dandapat, P.K. Bora. B-Scan Attentive CNN for the Classification of Retinal Optical Coherence Tomography Volumes. IEEE Signal Processing Letters. 27: 1025-1029, 2020.
134. M. Klimont, M. Flieger, J. Rzeszutek, J. Stachera, A. Zakrzewska, K. Jończyk-Potoczna. Automated Ventricular System Segmentation in Paediatric Patients

Treated for Hydrocephalus Using Deep Learning Methods. BioMed Research International. 3059170, 2019.
135. S.S. Mishra, B. Mandal, N.B. Puhan. Multi-Level Dual-Attention Based CNN for Macular Optical Coherence Tomography Classification. IEEE Signal Processing Letters. 26(12):1793-1797, 2019.
136. X. Dong, T. Wang, Y. Lei, et al. Synthetic CT generation from non-attenuation corrected PET images for Whole-body PET imaging. Physics in Medicine and Biology. 64(21):215016, 2019.
137. P. Song, Y.C. Eldar, G. Mazor, M.R.D. Rodrigues. HYDRA: Hybrid deep magnetic resonance fingerprinting. Medical Physics. 46(11):4951-4969, 2019.
138. T. Wang, Y. Lei, Z. Tian, et al. Deep learning-based image quality improvement for low-dose computed tomography simulation in radiation therapy. Journal of Medical Imaging. 6(4):43504, 2019.
139. Y. Wu, Y. Ma, J. Liu, J. Du, L. Xing. Self-attention convolutional neural network for improved MR image reconstruction. Information Sciences. 490:317-328, 2019.
140. A. Vaswani, N. Shazeer, N. Parmar, et al. Attention is all you need. NIPS; 2017.
141. Z. Liu, H. Mao, C.H. Wu, et al. A ConvNet for the 2020s. https://arxiv.org/pdf/2201.03545.pdf, 2022.
142. A. Chartsias, G. Papanastasiou, C. Wang, et al. Disentangle, Align and Fuse for Multimodal and Semi-Supervised Image Segmentation. IEEE Transactions on Medical Imaging. 40(3): 781-792, 2021.
143. A. Chartsias, T. Joyce, G. Papanastasiou, et al. Factorised Spatial Representation Learning: Application in Semi-supervised Myocardial Segmentation. In: Frangi, A., Schnabel, J., Davatzikos, C., Alberola-López, C., Fichtinger, G. (eds) MICCAI 2018. Lecture Notes in Computer Science: 11071, 2018.
144. A. Chartsias, G. Papanastasiou, C. Wang, et al. Multimodal cardiac segmentation using disentangled representations. Lecture Notes in Computer Science. In: MICCAI (STACOM Workshop) 2019; 12009, 2019.
145. H. Jiang, A. Chartsias, X. Zhang, et al. Semi-supervised Pathology Segmentation with Disentangled Representations. Lecture Notes in Computer Science. In: MICCAI (DART DCL) 2020; 12444, 2020.
146. A. Chartsias, T. Joyce, G. Papanastasiou, et al. Disentangled representation learning in cardiac image analysis. Medical Image Analysis. 58: 101535, 2019.
147. O. Ronneberger, P. Fischer, T. Brox. U-net: Convolutional networks for biomedical image segmentation. in International Conference on Medical image computing and computer-assisted intervention. Springer. 234–241, 2015.
148. O. Oktay, J. Schlemper, L.L. Folgoc, et al. Attention u-net: Learning where to look for the pancreas. arXiv preprint arXiv:1804.03999, 2018.
149. X. Xing, G. Papanastasiou, S. Walsh, G. Yang. Less is More: Unsupervised Mask-guided Annotated CT Image Synthesis with Minimum Manual Segmentations. IEEE Transactions on Medical Imaging, doi: 10.1109/TMI.2023.3260169, 2023.
150. G. Papanastasiou, M.C. Williams, L.E. Kershaw, et al. Measurement of myocardial blood flow by cardiovascular magnetic resonance perfusion: comparison of distributed parameter and Fermi models with single and dual bolus. J Cardiovasc Magn Reson. 17(1):17, 2015.
151. G. Papanastasiou, M.C. Williams, M.R. Dweck, et al. Quantitative assessment of myocardial blood flow in coronary artery disease by cardiovascular magnetic resonance: comparison of Fermi and distributed parameter modeling against invasive methods. J Cardiovasc Magn Reason. 18: 57, 2016.
152. V. Hamy, N. Dikaios, S. Punwani, et al. Respiratory motion correction in dynamic MRI using robust data decomposition registration-Application to DCE-MRI. Medical Image Analysis. 18(2): 301-313, 2014.
153. S. Bakas, et al. Identifying the best machine learning algorithms for brain tumor segmentation, progression assessment, and overall survival prediction in the BRATS challenge. arXiv preprint arXiv:1811.02629. https://arxiv.org/abs/1811.02629, 2018.
154. L. Csincsik, T.J. MacGillivray, E. Flynn, et al. Peripheral Retinal Imaging Biomarkers for Alzheimer’s Disease: A Pilot Study. Ophthalmic Res. 59 (4):182–19, 2018.
155. A. Villaplana-Velasco, M. Pigeyre, J. Engelmann. et al. Fine-mapping of retinal vascular complexity loci identifies Notch regulation as a shared mechanism with myocardial infarction outcomes. Commun Biol. 6: 523, 2023.
156. S.J. Wiseman, J.F. Zhang, C. Gray, et al. Retinal capillary microvessel morphology changes are associated with vascular damage and dysfunction in cerebral small vessel disease. Journal of Cerebral Blood Flow & Metabolism. 43(2):231-240, 2023.
157. G. Spyretta, D.D. Cokkinos. Recent advances in vascular ultrasound imaging technology and their clinical implications. Ultrasonics. 119: 106599, 2022.
158. S.A. Tsaftaris, X. Zhou, R. Tang, D. Li, R. Dharmakumar. Detecting myocardial ischemia at rest with cardiac phase-resolved blood oxygen level-dependent cardiovascular magnetic resonance. Circ Cardiovasc Imaging. 6(2): 311-9, 2013.
159. G. Papanastasiou, M.A. Rodrigues, C. Wang, et al. Pharmacokinetic modelling for the simultaneous assessment of perfusion and 18F-flutemetamol uptake in cerebral amyloid angiopathy using a reduced PET-MR acquisition time: Proof of concept. NeuroImage. 225: 117482, 2021.
160. G. Papanastasiou, M.C. Williams, M.R. Dweck, et al. Multimodality Quantitative Assessments of Myocardial Perfusion Using Dynamic Contrast Enhanced Magnetic Resonance and 15O-Labeled Water Positron Emission Tomography Imaging. in IEEE Transactions on Radiation and Plasma Medical Sciences. 2(3): 259-271, 2018.
161. C. Wang, G. Yang, G. Papanastasiou, et al. DiCyc: GAN-based deformation invariant cross-domain information fusion for medical image synthesis. Information Fusion. 67: 147-160, 2021.
162. C. Wang, G. Yang, G. Papanastasiou. Unsupervised Image Registration towards Enhancing Performance and Explainability in Cardiac and Brain Image Analysis. Sensors. 22(6):2125, 2022.
163. C. Wang, G. Yang, G. Papanastasiou. FIRE: Unsupervised bi-directional inter- and intra-modality registration using deep networks. 2021 IEEE 34th International Symposium on Computer-Based Medical Systems (CBMS), Aveiro, Portugal. 510-514. doi: 10.1109/CBMS52027.2021.00101, 2021.
164. B. Schölkopf, F. Locatello, S. Bauer, et al. Towards Causal Representation Learning. https://arxiv.org/pdf/2102.11107.pdf, 2021.
165. G. Zhou, L. Yao, X. Xu, et al. On the Opportunity of Causal Deep Generative Models: A Survey and Future Directions. https://arxiv.org/abs/2301.12351, 2023.
166. F.B. Ahmad, J.A. Cisewski, J. Xu, R.N. Anderson. Provisional Mortality Data — United States, 2022. MMWR Morb Mortal Wkly Rep. 72:488–492, 2023.
167. Radford A, Kim JW, Hallacy C, et al. Learning Transferable Visual Models From Natural Language Supervision. https://arxiv.org/pdf/2103.00020. pdf.2021.
168. ChatGPT: Optimizing Language Models for Dialogue. OpenAI. https://openai.com/blog/chatgpt/. Published November 30, 2022.
169. Brown T, Mann B, Ryder N, et al. Language models are few-shot learners. Advances in Neural Information Processing Systems 2020;33:1877–1901.
170. OpenAI. GPT-4 Technical report. https://arxiv.org/pdf/2303.08774.pdf. 2023.
171. Wang S, Zhao Z, Ouyang X, Wang Q, Shen D. (2023). Chatcad: Interactive computer-aided diagnosis on medical image using large language models. arXiv preprint arXiv:2302.07257. 2023.
172. Chen J, Guo H, Yi K, et al. Visualgpt: Data-efficient adaptation of pretrained language models for image captioning. Proceedings of the IEEE/CVF Conference on Computer Vision and Pattern Recognition. 2022.
173. Jeblick K, Schachtner B, Dexl J, Mittermeier A, Stüber AT, Topalis J, Weber T, Wesp P, Sabel B, Ricke J. ChatGPT Makes Medicine Easy to Swallow: An Exploratory Case Study on Simplified Radiology Reports. arXiv preprint arXiv:2212.14882. 2022.
174. Xiao Y, Wang WY. On Hallucination and Predictive Uncertainty in Conditional Language Generation. In: Proceedings of the 16th Conference of the European Chapter of the Association for Computational Linguistics: Main Volume. 2021.